\documentclass[11pt]{article}

\usepackage[final]{acl}

\usepackage{times}
\usepackage{latexsym}
\usepackage{tabularx}
\usepackage[T1]{fontenc}

\usepackage[utf8]{inputenc}

\usepackage{microtype}
\usepackage[table]{xcolor}
\usepackage{multirow}

\usepackage{inconsolata}

\usepackage{graphicx}
\usepackage{booktabs} 
\usepackage{multirow}
\usepackage{float}

\usepackage{subcaption}
\usepackage{listings}
\usepackage{pifont}

\definecolor{codegreen}{rgb}{0,0.6,0}
\definecolor{codegray}{rgb}{0.5,0.5,0.5}
\definecolor{codepurple}{rgb}{0.58,0,0.82}
\definecolor{backcolour}{rgb}{0.95,0.95,0.92}
\definecolor{mylightblue}{rgb}{0.68, 0.85, 0.9}
\definecolor{mywheat}{rgb}{0.96, 0.87, 0.7}
\lstdefinestyle{pythonstyle}{
    backgroundcolor=\color{backcolour},   
    commentstyle=\color{codegreen},
    keywordstyle=\color{magenta},
    numberstyle=\tiny\color{codegray},
    stringstyle=\color{codepurple},
    basicstyle=\ttfamily\footnotesize\tiny,
    breakatwhitespace=false,         
    breaklines=true,                 
    captionpos=b,                    
    keepspaces=true,                 
    numbers=left,                    
    numbersep=5pt,                  
    showspaces=false,                
    showstringspaces=false,
    showtabs=false,                  
    tabsize=2
}

\lstdefinestyle{prompt_style}{
    frame=single,
    basicstyle=\ttfamily\scriptsize,
    backgroundcolor=\color{white},
    stringstyle=\color{black},
    commentstyle=\color{darkgreen}\slshape,
    stringstyle=\color{darkred},
    numberstyle=\tiny\color{codegray},
    emphstyle=\color{pink}\underbar,
    breakindent=0pt,
    escapeinside={(*@}{@*)},
    breakatwhitespace=true,
    breaklines=true,
    captionpos=b,
    keepspaces=true,
    numbersep=5pt,
    showspaces=false,                
    showstringspaces=false,
    showtabs=false,
    tabsize=2,
}
\title{EduEval: A Hierarchical Cognitive Benchmark for Evaluating Large Language Models in Chinese Education}

\author{
\textbf{Guoqing Ma}\textsuperscript{1}, 
\textbf{Jia Zhu}\textsuperscript{1}, 
\textbf{Hanghui Guo}\textsuperscript{1}, 
\textbf{Weijie Shi}\textsuperscript{2}, 
\textbf{Yue Cui}\textsuperscript{2}, 
\textbf{Jiawei Shen}\textsuperscript{1}, 
\textbf{Zilong Li}\textsuperscript{1}, 
\textbf{Yidan Liang}\textsuperscript{1}\\[0.5em] 
\textsuperscript{1}Zhejiang Normal University, Zhejiang, China\\
\textsuperscript{2}Hong Kong University of Science and Technology, Hong Kong, China
}

\begin{document}
\maketitle
\begin{abstract}
Large language models (LLMs) demonstrate significant potential for educational applications. However, their unscrutinized deployment poses risks to educational standards, underscoring the need for rigorous evaluation. We introduce EduEval, a comprehensive hierarchical benchmark for evaluating LLMs in Chinese K-12 education. This benchmark makes three key contributions: (1) Cognitive Framework: We propose the EduAbility Taxonomy, which unifies Bloom’s Taxonomy and Webb’s Depth of Knowledge to organize tasks across six cognitive dimensions including Memorization, Understanding, Application, Reasoning, Creativity, and Ethics. (2) Authenticity: Our benchmark integrates real exam questions, classroom conversation, student essays, and expert-designed prompts to reflect genuine educational challenges; (3) Scale: EduEval comprises 24 distinct task types with over 11,000 questions spanning primary to high school levels. We evaluate 14 leading LLMs under both zero-shot and few-shot settings, revealing that while models perform well on factual tasks, they struggle with classroom dialogue classification and exhibit inconsistent results in creative content generation. Interestingly, several open source models outperform proprietary systems on complex educational reasoning. Few-shot prompting shows varying effectiveness across cognitive dimensions, suggesting that different educational objectives require tailored approaches. These findings provide targeted benchmarking metrics for developing LLMs specifically optimized for diverse Chinese educational tasks. The EduEval dataset and code are publicly available at \url{https://github.com/Maerzs/E_edueval}.

\end{abstract}

\section{Introduction}

The rapid advancement of large language models (LLMs) in natural language processing (NLP) and artificial intelligence (AI) has opened new possibilities for their application in education. With strong capabilities in language understanding, reasoning, and text generation, LLMs have been widely adopted in intelligent tutoring, automated grading, and educational content generation, significantly enhancing personalized learning and instructional efficiency \cite{brown2020language, ouyang2022training, wei2022chain}. Their impact now extends from primary and secondary education to higher education and vocational training \cite{glaser2001knowing, kasneci2023chatgpt, peng2023instruction}, with great potential shown in real world teaching activities like lesson preparation assistance and interactive classroom support \cite{kasneci2023chatgpt, chang2024survey, gan2023large, huang2025design}.

However, evaluating LLMs in authentic educational settings remains a significant challenge \cite{guo2021ai}. The educational process is inherently dynamic and heterogeneous \cite{koopmans2020education}, with diverse student cognitive profiles requiring constant adaptation \cite{haelermans2022effects}. Current mainstream evaluation methods rely heavily on static assessments \cite{chen2024gaia}, such as standardized multiple choice questions, which fail to capture the complex reasoning, generative expression, and ethical decision making required in real world classrooms \cite{allen2024ed, yan2024practical}. As a result, these evaluations tend to underestimate the true educational potential of LLMs \cite{srivastava2022beyond}. While general purpose benchmarks such as MMLU \cite{hendrycks2020measuring}, BIG-bench \cite{srivastava2022beyond}, and HELM \cite{liang2022holistic} cover broad abilities, they fall short in addressing the specific needs of educational applications. Similarly, Chinese benchmarks like E-Eval \cite{hou2024eval}, AGIEval \cite{zhong2023agieval}, and C-Eval \cite{huang2023c} are still dominated by low level cognitive questions and single-turn formats, lacking systematic evaluation of higher-order abilities and providing limited diversity in task contexts and cognitive depth \cite{long2024evaluating}.

To bridge this gap, a robust benchmark must assess a spectrum of cognitive abilities grounded in authentic pedagogical contexts. Established frameworks like Bloom’s Taxonomy of Educational Objectives \cite{krathwohl2002revision} and Webb’s Depth of Knowledge (DOK) model \cite{masharipova2024bloom} provide a foundation for structuring cognitive levels, but a comprehensive evaluation requires a synthesized framework tailored specifically for the multifaceted demands of K-12 education. We introduce \textbf{EduEval}, a hierarchical benchmark designed to meet this need by systematically evaluating LLMs on Chinese K-12 educational tasks. 

EduEval prioritizes educational authenticity by incorporating tasks directly from teaching practice. Unlike traditional benchmarks, it includes classroom dialogue classification, student essay scoring, and teaching scenario design activities that reflect educators' daily challenges. To ensure a comprehensive and authentic assessment, we developed a curriculum-balanced dataset spanning core subjects across developmental stages from primary to high school. All tasks derive from genuine educational sources, including examination archives, classroom transcripts, and expert-created materials. Our rigorous quality control employs a two-phase annotation protocol where practicing educators provide initial assessments before domain experts conduct thorough reviews, ensuring alignment with actual pedagogical standards rather than artificial benchmarking metrics.

Our systematic evaluation of mainstream language models (including both general-purpose and education-specific systems) reveals a consistent performance gradient across cognitive dimensions. Models generally excel at knowledge-oriented tasks, with top performers achieving over 90\% accuracy in areas like formula recall, but struggle significantly with contextual application in authentic educational scenarios, where accuracy can drop below 30\% for tasks such as classroom dialogue classification. This pattern mirrors the complexity progression in human learning and highlights key areas for model improvement. By identifying these capability patterns across the EduAbility hierarchy, our benchmark provides not merely a scoring system but a developmental blueprint for creating AI systems that can meaningfully support the full spectrum of educational processes.

In summary, we make three key contributions: 1) We propose the \textbf{EduAbility Taxonomy, a novel and hierarchical cognitive framework} that integrates Bloom’s Taxonomy and Webb’s Depth of Knowledge to systematically organize educational evaluation tasks across six cognitive dimensions. 2) We establish \textbf{EduEval, a comprehensive and authentic benchmark for Chinese K-12 education}, featuring over 11,000 questions across 24 task types derived from genuine classroom scenarios, including dialogue classification and essay scoring. 3) We present a \textbf{holistic evaluation of 14 leading LLMs} on this benchmark, providing a detailed analysis of their educational capabilities and a developmental blueprint for creating more effective AI systems for education.
\begin{figure*}
  \centering
    \includegraphics[width=\textwidth]{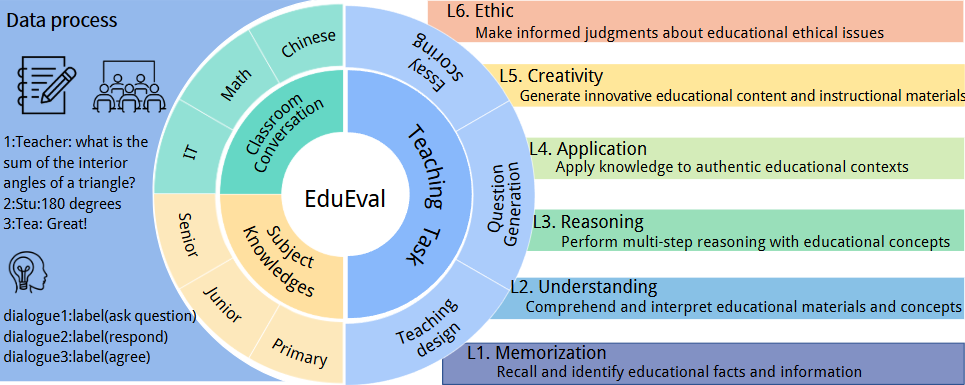}
  \caption{ Overview of EduEval: concentric rings represent data sources, with data processing methodology (left) and educational cognitive framework hierarchy (right).}
  \label{fig:kuangjia}
\end{figure*}
\section{Related Work}

In recent years, large language models (LLMs) have demonstrated remarkable capabilities across various domains through mechanisms such as pre-training, supervised fine-tuning, and alignment with human or AI feedback \cite{bai2022training,christiano2017deep,dong2023i3,dong2024unsupervised}. With the continuous advancement of natural language processing technology, benchmarks have expanded to encompass more complex tasks, including machine translation \cite{bojar2014findings} and summarization \cite{narayan2018don, hermann2015teaching}. Comprehensive benchmarks like GLUE \cite{wang2018glue} and SuperGLUE \cite{sarlin2020superglue} amalgamate various natural language understanding tasks, providing standards for evaluating model performance. Popular frameworks like MMLU \cite{hendrycks2020measuring} include tasks ranging from real-world exams to book knowledge, assessing ability in language understanding, common sense reasoning \cite{clark2018think,talmor2018commonsenseqa,sakaguchi2021winogrande}, and code generation \cite{chen2021evaluating,austin2021program}.

With the burgeoning development of Chinese Large Models, an increasing number of Chinese benchmarks have emerged. CLUE \cite{xu2020clue}, an influential Chinese NLU benchmark, has been widely adopted. In addition, Chinese benchmarks similar to MMLU have surfaced, such as MMCU \cite{zeng2023measuring}, which encompasses medicine and education within its four main domains. AGIEval \cite{zhong2023agieval} focuses on standardized Chinese exams, while C-Eval \cite{huang2023c} includes 13,948 questions on four levels of difficulty, from middle school to professional tests. M3KE \cite{liu2023m3ke} gathers 71 tasks from the Chinese education examination system, akin to the coverage of C-Eval. CMMLU \cite{li2023cmmlu}, designed for the Chinese language and cultural context, is a fully localized Chinese benchmark. These benchmarks mainly focus on low-level cognitive questions and single-turn formats, lacking systematic assessment of higher-order abilities and offering limited task context diversity and cognitive depth.

EduEval actualizes this logical progression through its hierarchical cognitive framework, the EduAbility Taxonomy. This structure is what enables the benchmark to incorporate authentic assessments derived from real classroom materials and, for the first time, systematically evaluate the creative, critical, and ethical dimensions that were fundamentally beyond the scope of prior work.

\section{EduEval }

\subsection{Design Principles}
Traditional cognitive taxonomies present limitations when applied to comprehensive LLM evaluation in educational contexts. Bloom's Taxonomy \cite{krathwohl2002revision}, while influential, assumes rigid hierarchical progression and lacks consideration of concurrent cognitive processes. Webb's Depth of Knowledge (DOK) model \cite{masharipova2024bloom} addresses cognitive complexity but inadequately captures the multidimensional nature of educational cognition. Neither framework sufficiently addresses ethical reasoning, which is crucial for educational AI systems. We propose the EduAbility Taxonomy to address these limitations by integrating both frameworks' strengths while introducing novel dimensions. Our approach maps Memory and Understanding levels to DOK Levels 1-2 and Bloom's foundational skills, Application and Reasoning levels to DOK Level 3 with Bloom's analytical capabilities, and Creativity level to Bloom's highest-order thinking and DOK Level 4 demands. Additionally, we introduce an Ethics dimension to address moral reasoning in educational contexts.
\begin{table*}[t]
\centering
\small
\caption{EduEval task distribution across six cognitive levels (24 tasks, 11,150 instances).}
\label{tab:task-compact}
\renewcommand{\arraystretch}{1}
\begin{tabular}{p{0.38\textwidth} >{\centering\arraybackslash}p{0.08\textwidth} p{0.38\textwidth} >{\centering\arraybackslash}p{0.08\textwidth}}
\toprule 
\textbf{Task (ID)} & \textbf{Metric} & \textbf{Task (ID)} & \textbf{Metric} \\
\midrule 
\rowcolor{gray!20}
\multicolumn{1}{l}{\textbf{Memory}} & & \multicolumn{1}{l}{\textbf{Understanding}} & \\
\midrule
Primary Formula Recall (1-1) & Acc. & Primary Understanding (2-1) & Acc. \\
Junior Knowledge Recall (1-2) & Acc. & Junior Understanding (2-2) & Acc. \\
Senior Concept Recall (1-3) & Acc. & Senior Understanding (2-3) & Acc. \\
 & & Poetry Appreciation (2-4) & Rouge-L \\
 & & Reading Comprehension (2-5) & Rouge-L \\
\midrule
\rowcolor{gray!20}
\multicolumn{1}{l}{\textbf{Application}} & & \multicolumn{1}{l}{\textbf{Reasoning}} & \\
\midrule
Classroom Dialogue (3-1) & Acc. & General Logical Inference (4-1) & Acc. \\
Primary Problem Solving (3-2) & Acc. & Primary Reasoning (4-2) & Acc. \\
Junior Problem Solving (3-3) & Acc. & Junior Reasoning (4-3) & Acc. \\
Senior Problem Solving (3-4) & Acc. & Senior Reasoning (4-4) & Acc. \\
Essay Scoring (3-5) & RMSE & & \\
\midrule
\rowcolor{gray!20}
\multicolumn{1}{l}{\textbf{Creativity}} & & \multicolumn{1}{l}{\textbf{Ethics}} & \\
\midrule
Question Generation (5-1) & GPT & Primary Moral (6-1) & Acc. \\
Teaching Design (5-2) & GPT & Junior Ethics Scenario (6-2) & Acc. \\
Text Writing (5-3) & GPT & Senior Ethics Scenario (6-3) & Acc. \\
 & & Educational Ethics (6-4) & Acc. \\
\bottomrule 
\end{tabular}
\begin{flushleft}
\vspace{1pt} 
\footnotesize
\textit{Note:} Acc. = Accuracy; GPT = GPT-based evaluation; RMSE = Root Mean Square Error (lower is better). Full details in table \ref{tab:edueval-tasks}.
\end{flushleft}
\end{table*}

As depicted in Figure \ref{fig:kuangjia}, the EduAbility Taxonomy organizes tasks into six hierarchical cognitive levels: Memory, Understanding, Application, Reasoning, Creativity, and Ethics. These levels form an interconnected cognitive scaffold where higher-level tasks build on foundational abilities while operating synergistically. Memory tasks assess recall of core concepts. Understanding requires interpretation and explanation. Application involves real world problem-solving (e.g., Essay Scoring, Classroom Dialogue Classification), Reasoning evaluates causal analysis and complex inference. Creativity encompasses innovative generation and knowledge integration, and Ethics addresses moral judgment in educational contexts. The theoretical foundations, detailed framework mappings, and empirical validation of this taxonomy are extensively discussed in the Appendix \ref{appendix:eduability_framework}.

Each level includes several specific evaluation tasks that correspond to the respective abilities. For example, our Classroom Dialogue Classification tasks utilize real teacher-student interactions, while our Essay Scoring task requires evaluating student compositions according to standard educational rubrics. It is essential to note that this taxonomy does not imply a linear progression of abilities: the cognitive dimensions are interdependent, with higher-level reasoning often requiring strong foundational knowledge, and creativity building upon solid understanding. By synthesizing performance across multiple tasks within each level, EduEval enables precise identification of where models excel versus where they struggle, providing a framework for targeted improvements in capabilities most relevant to educational applications. As this taxonomy continues to evolve, it will increasingly facilitate comprehensive evaluation of LLMs across the full spectrum of cognitive dimensions essential for effective educational support from elementary through high school settings.

\subsection{Data Collection and Processing}
\textbf{Data Source.}
EduEval is constructed from a blend of authentic educational content and expert-authored materials to comprehensively evaluate language models across six cognitive levels. For memory, understanding, reasoning, and application tasks, we collect standardized exam questions from Gaokao, Zhongkao, and various primary and middle school assessments, covering core subjects such as Chinese, Mathematics, English, and Science. The essay scoring task is based on real student-written essays collected from diverse school contexts, each paired with ground-truth scores assigned by certified instructors using official grading rubrics. Classroom dialogue classification tasks are built from transcripts of real classroom interactions, which are annotated by experienced educators according to a predefined taxonomy of pedagogical discourse types. Creative tasks and ethical reasoning tasks are entirely designed by domain experts. While creative tasks such as teaching design or question generation require open-ended model responses, ethical tasks simulate realistic dilemmas related to fairness, bias, or privacy in school settings, and are presented in structured formats with predefined choices. These tasks are intended to probe the model’s moral and normative alignment, without relying on externally annotated labels. A detailed mapping between task types and their corresponding data sources is provided in Appendix~\ref{appendix:data_source}.

\textbf{Data Processing.}
To operationalize the EduAbility cognitive scaffold on authentic and often unstructured educational materials, a methodological architecture of commensurate sophistication is required. Consequently, we designed and implemented a multi-agent processing pipeline whose structure is intentionally engineered to mirror the interconnected nature of the cognitive framework itself. This architecture serves the dual purpose of ensuring high-fidelity conversion of diverse raw materials (PDFs, Word documents, web content) into a unified JSON schema, while systematically embedding our theoretical principles into the data curation process. The pipeline includes: (1) Format Conversion Agent for OCR extraction and mathematical notation processing; (2) Standardization Agent for schema validation and field extraction; (3) Task Format Optimizer that renders objective tasks as multiple-choice and preserves free-form structure for open ended tasks; (4) Quality Control Agent for duplicate detection and anomaly flagging. This architecture enables scalable processing of 11,150+ items while maintaining high fidelity. Details in Appendix~\ref{appendix:data-processing}.

\textbf{Data Annotation.}
The annotation process itself is a hybrid human-AI framework, engineered to embed our theoretical principles directly into the dataset. A Task Routing Agent first assigns items to appropriate annotation tracks based on task type, after which Annotation Support Agents provide draft suggestions that human experts can accept or override. All subjective labels then undergo a rigorous double-blind cross-review, managed by a Consensus Orchestration Agent, to achieve an inter-annotator agreement threshold of over 90\%. Crucially, a Cognitive Alignment Verification Agent operationalizes our theoretical framework, using a fine-tuned classifier to validate that each item is rigorously mapped to its primary dimension within the EduAbility cognitive scaffold. This multi-agent, human-in-the-loop approach ensures that each benchmark item is not only internally consistent but also theoretically grounded, combining the irreplaceable judgment of educational experts with the systematic rigor of AI-driven validation. Complete protocols in Appendix~\ref{appendix:expert-annotation}.

\begin{table*}[h!]
\scriptsize
\renewcommand{\arraystretch}{1}
\centering
\caption{Zero-shot performance (\%) of various models on Memorization, Understanding, and Application levels. Different depth of color of the cells indicate that the values inside is higher. $^*$ Tasks 2-4 and 2-5 are evaluated using Rouge-L metric; $^\dagger$ task 3-5 uses RMSE metric (lower is better); all other tasks are evaluated using accuracy.}
\label{tab:zero-shot-mua}
\resizebox{\textwidth}{!}{
\begin{tabular}{l|ccc|ccccc|ccccc}
\hline
\multirow{2}{*}{\textbf{Model}} 
& \multicolumn{3}{c|}{\textbf{Memory}} 
& \multicolumn{5}{c|}{\textbf{Understanding}} 
& \multicolumn{5}{c}{\textbf{Application}} \\
& 1-1 & 1-2 & 1-3 
& 2-1 & 2-2 & 2-3 & 2-4$^*$ & 2-5$^*$ 
& 3-1 & 3-2 & 3-3 & 3-4 & 3-5$^\dagger$ \\
\hline
Spark-X1 & \cellcolor{orange!60}92.5 & \cellcolor{orange!60}86.2 & \cellcolor{orange!60}91.0 & \cellcolor{orange!60}93.1 & \cellcolor{orange!60}86.6 & \cellcolor{orange!30}88.4 & 48.9 & 50.5 & 27.8 & \cellcolor{orange!60}95.4 & \cellcolor{orange!60}87.6 & \cellcolor{orange!60}91.0 & 43.175 \\
Qwen-plus & \cellcolor{orange!30}81.8 & \cellcolor{orange!30}82.0 & 89.2 & \cellcolor{orange!30}89.4 & \cellcolor{orange!30}85.3 & \cellcolor{orange!60}90.9 & 47.7 & 50.6 & \cellcolor{orange!30}24.8 & 75.9 & 83.2 & 79.0 & \cellcolor{orange!30}12.240 \\
Qwen-72B & 79.8 & 78.8 & \cellcolor{orange!30}89.3 & 85.7 & 81.4 & 86.8 & 48.4 & 50.1 & \cellcolor{orange!30}24.8 & 74.4 & 80.9 & 72.1 & 32.815 \\
DeepseekR1-32B & 78.5 & 80.0 & 89.0 & 86.8 & 83.4 & 82.0 & 47.8 & 50.7 & 18.1 & \cellcolor{orange!30}79.3 & \cellcolor{orange!30}84.8 & \cellcolor{orange!30}83.4 & 21.410 \\
GPT-4o & 72.0 & 74.9 & 81.2 & 82.2 & 81.3 & 80.7 & 48.5 & \cellcolor{orange!30}51.3 & \cellcolor{orange!60}28.2 & 58.9 & 76.6 & 58.3 & \cellcolor{orange!60}9.013 \\
Yi-34B & 74.3 & 70.9 & 79.8 & 79.0 & 78.0 & 80.2 & 49.3 & \cellcolor{orange!30}51.3 & 17.3 & 60.8 & 77.0 & 60.8 & 28.528 \\
Qwen-14B-Chat & 71.0 & 73.0 & 78.8 & 81.9 & 77.6 & 78.2 & \cellcolor{orange!60}49.8 & \cellcolor{orange!60}51.8 & 17.1 & 50.2 & 76.6 & 54.8 & 34.109 \\
GLM4-9B-Chat & 67.8 & 72.4 & 77.4 & 77.4 & 78.8 & 80.7 & \cellcolor{orange!30}49.7 & 50.8 & 22.2 & 52.8 & 73.6 & 61.0 & 28.260 \\
Qwen-7B & 60.0 & 62.3 & 63.0 & 68.7 & 68.4 & 68.9 & 48.8 & \cellcolor{orange!30}51.3 & 15.5 & 44.3 & 66.7 & 49.5 & 35.814 \\
Yi-6B & 56.8 & 61.0 & 60.4 & 62.2 & 62.5 & 61.4 & 47.8 & 51.1 & 14.9 & 41.7 & 63.4 & 47.5 & 35.742 \\
EduChat-sft-002-13b & 51.1 & 51.2 & 52.8 & 59.7 & 58.7 & 59.1 & 48.9 & 49.9 & 20.5 & 35.7 & 56.8 & 39.9 & 13.3 \\
LLaMA3-8B & 41.0 & 43.8 & 44.3 & 48.1 & 47.5 & 47.8 & 42.4 & 46.2 & 10.1 & 32.4 & 48.5 & 34.1 & 12.476 \\
BaiChuan-13b & 39.1 & 45.9 & 39.9 & 49.4 & 48.1 & 46.1 & 49.1 & 50.1 & 11.5 & 33.9 & 48.3 & 38.2 & 33.393 \\
LLaMA2-Chinese-13B & 28.1 & 34.2 & 32.5 & 32.5 & 32.2 & 33.0 & 47.1 & 50.8 & 16.4 & 27.7 & 34.6 & 26.2 & 26.298 \\
\hline
\end{tabular}
}
\end{table*}

\begin{table*}[h!]
\scriptsize
\renewcommand{\arraystretch}{1}
\centering
\caption{Zero-shot performance (\%) of various models on Reasoning, Creation, and Ethics levels. Different depth of color of the cells indicate that the values inside is higher. All tasks are evaluated using accuracy except for $^*$ creativity tasks (5-1, 5-2, 5-3), which use GPT-based evaluation.}
\label{tab:zero-shot-rce}
\resizebox{\textwidth}{!}{
\begin{tabular}{l|cccc|ccc|cccc|cc}
\hline
\multirow{2}{*}{\textbf{Model}} & \multicolumn{4}{c|}{\textbf{Reasoning}} & \multicolumn{3}{c|}{\textbf{Creation}$^*$} & \multicolumn{4}{c|}{\textbf{Ethics}} & \multirow{2}{*}{\textbf{Average}} & \multirow{2}{*}{\textbf{Rank}} \\
& 4-1 & 4-2 & 4-3 & 4-4 & 5-1 & 5-2 & 5-3 & 6-1 & 6-2 & 6-3 & 6-4 & & \\
\hline
Spark-X1 & \cellcolor{orange!60}78.8 & \cellcolor{orange!60}90.6 & \cellcolor{orange!60}93.6 & \cellcolor{orange!60}88.0 & \cellcolor{orange!30}88.3 & 85.9 & \cellcolor{orange!60}88.7 & 74.4 & 81.0 & 77.8 & 79.8 & \cellcolor{orange!60}81.1 & \cellcolor{orange!60}1 \\
Qwen-plus & \cellcolor{orange!30}72.6 & 69.2 & \cellcolor{orange!30}88.4 & \cellcolor{orange!30}85.8 & 86.3 & 87.0 & \cellcolor{orange!30}86.4 & \cellcolor{orange!30}80.0 & \cellcolor{orange!60}85.4 & \cellcolor{orange!60}83.4 & \cellcolor{orange!60}85.4 & \cellcolor{orange!30}77.7 & \cellcolor{orange!30}2 \\
Qwen-72B & 69.6 & 65.7 & 84.6 & 78.8 & 87.8 & \cellcolor{orange!60}90.4 & 80.0 & 78.7 & 79.3 & 78.2 & 78.3 & 75.0 & 3 \\
DeepseekR1-32B & 67.9 & \cellcolor{orange!30}75.4 & 66.7 & 78.6 & 85.5 & 86.3 & 81.1 & 76.8 & 79.2 & 78.8 & \cellcolor{orange!30}80.2 & 74.8 & 4 \\
GPT-4o & 65.4 & 54.7 & 71.8 & 71.6 & \cellcolor{orange!60}89.4 & 84.6 & 84.9 & 76.4 & 81.4 & 78.6 & 80.0 & 71.0 & 5 \\
Yi-34B & 58.1 & 56.0 & 78.8 & 66.0 & 83.9 & \cellcolor{orange!30}90.0 & 78.7 & 75.6 & 81.2 & 79.8 & 79.2 & 69.8 & 6 \\
Qwen-14B-Chat & 40.0 & 50.0 & 78.4 & 66.2 & 77.9 & 88.0 & 72.1 & \cellcolor{orange!60}84.0 & \cellcolor{orange!30}84.2 & \cellcolor{orange!30}81.6 & 80.0 & 68.1 & 7 \\
GLM4-9B-Chat & 59.4 & 46.8 & 76.6 & 68.0 & 77.0 & 86.5 & 74.8 & 72.0 & 79.2 & 74.6 & 74.4 & 67.6 & 8 \\
Qwen-7B & 38.4 & 44.8 & 66.8 & 60.6 & 80.7 & 87.8 & 71.4 & 69.6 & 72.6 & 71.6 & 73.8 & 60.8 & 9 \\
Yi-6B & 40.6 & 41.4 & 60.4 & 53.5 & 74.0 & 88.4 & 77.0 & 70.8 & 63.2 & 69.5 & 62.9 & 57.9 & 10 \\
EduChat-sft-002-13B & 31.8 & 36.5 & 49.5 & 50.9 & 71.2 & 84.6 & 67.5 & 69.6 & 78.2 & 73.2 & 73.4 & 53.8 & 11 \\
LLaMA3-8B & 39.0 & 34.1 & 39.1 & 43.1 & 85.5 & 86.4 & 77.7 & 70.0 & 68.8 & 69.0 & 65.0 & 50.3 & 12 \\
BaiChuan-13b & 35.3 & 29.7 & 41.0 & 42.3 & 70.0 & 85.8 & 66.6 & 62.0 & 61.8 & 64.2 & 61.8 & 48.7 & 13 \\
LLaMA2-Chinese-13B & 25.7 & 25.9 & 21.8 & 23.9 & 63.0 & 85.4 & 55.7 & 59.5 & 49.8 & 47.8 & 50.6 & 39.3 & 14 \\
\hline
\end{tabular}
}
\end{table*}

\subsection{Task Definitions}
To evaluate LLMs across varied cognitive dimensions, we construct tasks grounded in the EduAbility Taxonomy, covering memory, understanding, application, reasoning, creativity, and ethics. Memory tasks assess factual recall, such as mathematical formulas and historical facts. Understanding tasks test conceptual interpretation, including reading comprehension and poetry analysis. Application tasks reflect real-world classroom scenarios like essay scoring and dialogue classification. Reasoning tasks involve logical inference and cross-disciplinary integration. Creativity tasks require open-ended generation, such as teaching design and question creation. Ethics tasks present dilemmas on fairness and moral judgment in education. A summary of task categories is shown in Table \ref{tab:task-compact}, with further details and examples in Appendix \ref{appendix:task definition}.

\section{Evaluation}

\subsection{Experimental Setup}
We evaluate all LLMs in both zero-shot and few-shot settings. In the zero-shot setting, each input consists of a task instruction and query, while in the few-shot setting, we provide three illustrative examples per task. To ensure consistency, all prompts are presented in Chinese, and for chat-based models, we preserve their native dialogue format. We set the decoding temperature to 0 during inference to reduce randomness in model outputs. When an input exceeds the model’s maximum context length, we truncate it from the middle to preserve instruction and answer segments. Multiple-choice tasks are evaluated using Accuracy, and generative tasks are evaluated using Rouge-L or GPT-based alignment metrics. The evaluation metrics used for each task are listed in Table~\ref{tab:task-compact}, and prompt formats are provided in Appendix~\ref{prompt-templates}. For experimental hardware configuration, refer to Appendix \ref{appendix:experimental-setup}.

\subsection{Evaluated Models}
We evaluated 14 large language models varying in size, organization, and design objective, as summarized in Table~\ref{tab:evaluated-models}. For commercial models, we included GPT-4o \cite{hurst2024gpt}, Qwen-14B-Chat \cite{bai2023qwen}, GLM4-9B-Chat \cite{zeng2022glm}, and Spark-X1 \cite{iflytek2024spark}, representing leading proprietary systems. Among open-source systems, we assessed Qwen-72B and Qwen-7B \cite{bai2023qwen}, Yi-34B-Chat and Yi-6B-Chat \cite{01AI2023Yi}, Chinese-LLaMA2-13B \cite{cui2023efficient}, DeepSeek-R1-32B \cite{deepseek2024}, LLaMA3-8B \cite{grattafiori2024llama}, and Baichuan-13B \cite{yang2023baichuan}. These models span a wide range of parameter scales and alignment strategies, enabling controlled comparisons across reasoning, generation, and ethical tasks. To further explore domain-specific capabilities, we included EduChat-sft-002-13B-Baichuan \cite{dan2023educhat}, which is fine-tuned from Baichuan-13B specifically for educational tasks, allowing for a direct analysis of the impact of domain specialization. The relevant model details are in the Appendix \ref{appendix:experimental-setup}.

\subsection{Experimental Results}
\label{Experimental-results}
We present the zero‐shot performance of all 14 models in Tables~\ref{tab:zero-shot-mua} and~\ref{tab:zero-shot-rce}, focusing here on the leading systems to highlight key patterns. From these results, we draw the following interconnected insights:

\textbf{Clear performance hierarchy among models.} The proprietary Spark-X1 leads overall rankings, with the Qwen series following closely, while the closed-source GPT-4o ranks fifth. This demonstrates that well-optimized open-source models can now match proprietary systems on educational tasks. Scale effects are significant within the same family, larger models (such as Yi-34B compared to Yi-6B) consistently outperform on complex tasks, confirming that increased parameter count substantially enhances educational reasoning and problem-solving capabilities.
  
\textbf{Task difficulty gradient and capability distribution.} All models exhibit consistent performance decline as task complexity increases. Models excel at memory and comprehension tasks but face challenges with application and reasoning tasks. This pattern reveals that while current LLMs have matured in surface-level language understanding and knowledge retrieval, they still have room for improvement in tasks requiring multi-step logical deduction and subject knowledge application. Notably, models perform better at recalling complex high school knowledge than elementary basics, potentially reflecting the overrepresentation of advanced academic material in training corpora.

\begin{table*}[h!]
\scriptsize
\renewcommand{\arraystretch}{1}
\centering
\caption{Few-shot performance (\%) of various models on Memory, Understanding, and Application levels. Different depth of color of the cells indicate that the values inside is higher. $^*$ Tasks 2-4 and 2-5 use Rouge-L metric; $^\dagger$ task 3-5 uses RMSE metric (lower is better); all other tasks use accuracy. Arrows indicate performance change compared to zero-shot: ↑ shows improvement, ↓ for 3-5 indicates improvement (lower RMSE).}
\label{tab:few-shot-mua}
\resizebox{\textwidth}{!}{
\begin{tabular}{l|ccc|ccccc|ccccc}
\hline
\multirow{2}{*}{\textbf{Model}} & \multicolumn{3}{c|}{\textbf{Memory}} & \multicolumn{5}{c|}{\textbf{Understanding}} & \multicolumn{5}{c}{\textbf{Application}} \\
& 1-1 & 1-2 & 1-3 & 2-1 & 2-2 & 2-3 & 2-4$^*$ & 2-5$^*$ & 3-1 & 3-2 & 3-3 & 3-4 & 3-5$^\dagger$ \\
\hline
Spark-X1 & \cellcolor{orange!60}91.3 & \cellcolor{orange!60}85.2 & \cellcolor{orange!60}91.2 & \cellcolor{orange!60}92.7 & \cellcolor{orange!60}86.6 & \cellcolor{orange!30}89.8↑ & 48.6 & 50.4 & 26.6 & \cellcolor{orange!60}94.6 & \cellcolor{orange!60}85.2 & \cellcolor{orange!60}88.6 & 23.0↓ \\
Qwen-plus & \cellcolor{orange!30}82.2↑ & \cellcolor{orange!30}80.6 & \cellcolor{orange!30}89.2 & \cellcolor{orange!30}88.8 & \cellcolor{orange!30}84.0 & \cellcolor{orange!60}91.0↑ & 48.9↑ & 51.0↑ & \cellcolor{orange!30}27.4↑ & \cellcolor{orange!30}78.3↑ & \cellcolor{orange!30}83.8↑ & \cellcolor{orange!30}76.8 & \cellcolor{orange!60}8.7↓ \\
GPT-4o & 72.7↑ & 74.3 & 79.8 & 83.3↑ & 82.4↑ & 81.2↑ & \cellcolor{orange!30}49.2↑ & \cellcolor{orange!60}51.7↑ & \cellcolor{orange!60}29.1↑ & 62.7↑ & 76.8↑ & 58.7↑ & \cellcolor{orange!30}8.9↓ \\
DeepseekR1-32B & 79.3↑ & 60.0 & 86.6 & 85.9 & 53.1 & 84.6↑ & 47.5 & 50.5 & 21.4↑ & 68.0 & 52.8 & 75.8 & 20.6↓ \\
Qwen-72B & 78.3 & 75.6 & 88.6 & 84.9 & 79.0 & 85.4 & 47.6 & \cellcolor{orange!30}51.6↑ & 25.4↑ & 80.9↑ & 78.6 & 74.9↑ & 15.4↓ \\
Qwen-14B-Chat & 69.3 & 71.9 & 79.4↑ & 79.9 & 77.7↑ & 82.9↑ & \cellcolor{orange!60}49.8 & \cellcolor{orange!30}51.6 & 20.2↑ & 53.5↑ & 72.9 & 60.6↑ & 10.6↓ \\
GLM4-9B-Chat & 62.1 & 69.3 & 73.6 & 65.7 & 73.8 & 75.5 & 49.1 & 48.2 & 19.3 & 64.0↑ & 70.5 & 62.4↑ & 8.4↓ \\
Yi-34B & 66.3 & 65.5 & 71.3 & 77.5 & 75.0 & 76.4 & 48.8 & 50.7 & 18.9↑ & 54.5 & 71.1 & 54.5 & 9.2↓ \\
Qwen-7B & 54.9 & 59.0 & 55.6 & 65.7 & 63.9 & 64.1 & 48.7 & 51.2 & 13.2 & 39.9 & 62.2 & 45.8 & 13.8↓ \\
EduChat-sft-002-13b & 51.6↑ & 52.4↑ & 52.1 & 56.8 & 59.9↑ & 59.7↑ & 47.9 & 42.9 & 23.5↑ & 34.2 & 59.8↑ & 38.8 & 12.1↓ \\
Yi-6B & 48.8 & 56.9 & 48.3 & 52.2 & 50.4 & 41.3 & 48.7↑ & 50.8 & 12.9 & 50.8↑ & 50.4 & 43.5 & 29.3↓ \\
Baichuan-13B & 39.5↑ & 47.2↑ & 39.9 & 46.0 & 49.7↑ & 49.9↑ & 47.3 & 44.0 & 17.8↑ & 31.7 & 52.5↑ & 37.7 & 13.6↓ \\
LLaMA3-8B & 38.0 & 42.6 & 39.0 & 47.5 & 41.6 & 47.9↑ & 43.4↑ & 46.0 & 14.5↑ & 43.1↑ & 48.9↑ & 37.8↑ & 16.4↑ \\
LLaMA2-Chinese-13B & 35.0↑ & 34.7↑ & 29.0 & 32.0 & 29.8 & 31.9 & 47.4↑ & 50.4 & 11.6 & 20.6 & 30.6 & 26.0 & 21.2↓ \\
\hline
\end{tabular}
}
\end{table*}

\begin{table*}[h!]
\scriptsize
\renewcommand{\arraystretch}{1}
\centering
\caption{Few-shot performance (\%) of various models on Reasoning, Creativity, and Ethics levels. Different depth of color of the cells indicate that the values inside is higher. All tasks use accuracy except for $^*$ creativity tasks (5-1, 5-2, 5-3) which use GPT-based evaluation. Arrows indicate performance change compared to zero-shot.}
\label{tab:few-shot-rce}
\resizebox{\textwidth}{!}{
\begin{tabular}{l|cccc|ccc|cccc|cc}
\hline
\multirow{2}{*}{\textbf{Model}} & \multicolumn{4}{c|}{\textbf{Reasoning}} & \multicolumn{3}{c|}{\textbf{Creativity}$^*$} & \multicolumn{4}{c|}{\textbf{Ethics}} & \multirow{2}{*}{\textbf{Average}} & \multirow{2}{*}{\textbf{Rank}} \\
& 4-1 & 4-2 & 4-3 & 4-4 & 5-1 & 5-2 & 5-3 & 6-1 & 6-2 & 6-3 & 6-4 & & \\
\hline
Spark-X1 & \cellcolor{orange!60}77.8 & \cellcolor{orange!60}89.1 & \cellcolor{orange!60}94.2↑ & \cellcolor{orange!60}87.6 & 74.0 & 43.6 & 54.5 & 81.2↑ & 84.2↑ & 80.0↑ & 82.4↑ & \cellcolor{orange!60}77.4 & \cellcolor{orange!60}1 \\
Qwen-plus & \cellcolor{orange!30}70.8 & 69.7↑ & 87.2 & \cellcolor{orange!30}84.8 & 78.3 & \cellcolor{orange!60}59.0 & \cellcolor{orange!30}64.8 & 82.0↑ & \cellcolor{orange!60}89.4↑ & \cellcolor{orange!60}87.4↑ & \cellcolor{orange!60}87.4↑ & \cellcolor{orange!30}75.8 & \cellcolor{orange!30}2 \\
Qwen-72B & 69.4 & \cellcolor{orange!30}70.3↑ & 86.0↑ & 79.0↑ & \cellcolor{orange!30}79.7 & 35.9 & 62.2 & 79.6↑ & 82.1↑ & 80.6↑ & 81.8↑ & 72.1 & 3 \\
DeepseekR1-32B & 70.0↑ & 61.2 & \cellcolor{orange!30}88.1↑ & 77.9 & \cellcolor{orange!60}83.5 & \cellcolor{orange!30}52.4 & \cellcolor{orange!60}70.1 & 81.2↑ & 83.6↑ & \cellcolor{orange!30}82.0↑ & 83.4↑ & 69.5 & 4 \\
GPT-4o & 64.7 & 52.2 & 73.8↑ & 72.8↑ & 63.2 & 37.5 & 43.0 & \cellcolor{orange!60}83.2↑ & \cellcolor{orange!30}88.2↑ & 81.8↑ & \cellcolor{orange!30}86.0↑ & 67.3 & 5 \\
Qwen-14B-Chat & 51.0↑ & 53.0↑ & 77.4 & 74.5↑ & 71.0 & 39.4 & 45.7 & \cellcolor{orange!30}82.4 & 81.4 & 81.2 & 72.4 & 65.2 & 6 \\
GLM4-9B-Chat & 56.6 & 50.2↑ & 76.7↑ & 66.6 & 58.9 & 32.2 & 53.1 & 72.0 & 84.0↑ & 79.8↑ & 82.8↑ & 62.9 & 7 \\
Yi-34B & 52.8 & 47.8 & 66.8 & 62.7 & 63.3 & 34.5 & 41.2 & 77.6↑ & 81.0 & 80.6↑ & 81.2↑ & 61.7 & 8 \\
Qwen-7B & 38.4 & 40.9 & 62.2 & 56.3 & 62.6 & 40.7 & 29.8 & 67.6 & 69.4 & 75.4↑ & 67.0 & 53.7 & 9 \\
EduChat-sft-002-13B & 30.2 & 37.2↑ & 51.5↑ & 50.2 & 57.2 & 43.1 & 40.3 & 73.4↑ & 75.2 & 74.2↑ & 71.4 & 53.2 & 10 \\
Yi-6B & 39.5 & 42.2↑ & 51.4 & 46.3 & 66.7 & 39.5 & 36.5 & 73.6↑ & 78.6↑ & 83.4↑ & 81.2↑ & 52.3 & 11 \\
BaiChuan-13B & 33.5 & 30.7↑ & 42.2↑ & 41.3 & 56.9 & 41.7 & 38.1 & 68.8↑ & 60.0 & 67.5↑ & 58.6 & 45.9 & 12 \\
LLaMA3-8B & 37.7 & 35.2↑ & 40.4↑ & 38.2 & 42.8 & 20.6 & 36.4 & 63.2 & 70.2↑ & 76.6↑ & 58.6 & 43.5 & 13 \\
LLaMA2-Chinese-13B & 27.4↑ & 21.2 & 26.2↑ & 28.7↑ & 52.2 & 34.8 & 28.6 & 47.2 & 48.8 & 55.2↑ & 51.4↑ & 36.8 & 14 \\
\hline
\end{tabular}
}
\end{table*}

\textbf{Instruction tuning advantages and domain adaptation limitations.} The impact of domain-specific adaptation is revealing: the education-specialized EduChat-sft-002-13B substantially improved upon its Baichuan-13B base model in most authentic educational contexts, especially in application and ethics tasks. However, this targeted enhancement came with a subtle trade-off, as performance slightly decreased on tasks assessing abstract reasoning independent of educational context. This suggests that the fine-tuning process, while optimizing for educational alignment, may have shifted the model's learned distribution away from the general-purpose capabilities inherited from its foundation. This nuanced finding underscores a critical challenge: narrow domain specialization, while effective for contextual tasks, can inadvertently constrain broader reasoning, highlighting the need for careful balance during fine-tuning.
\textbf{Subject differences and cognitive preferences.} Models generally perform better on humanities-based tasks than on science applications, reflecting their natural advantage in language and text processing. In contrast, mathematical problem-solving and physics applications require stronger symbolic manipulation and logical reasoning abilities, areas where performance is relatively weaker. This disparity suggests that future model training should place greater emphasis on developing scientific thinking capabilities.
  
\textbf{Strong creative and ethical capabilities.} Leading models demonstrate impressive abilities in educational content creation, with multiple systems performing excellently on creative generation tasks. Models also perform well in ethical reasoning, with Qwen-plus showing particularly outstanding results across ethical dimensions. These findings indicate that current LLMs have developed significant proficiency in generating pedagogically valuable materials and navigating ethical considerations in educational contexts.

\textbf{Does Few-shot prompting help?}Tables \ref{tab:few-shot-mua} and Table \ref{tab:few-shot-rce} show the few-shot performance of various models across six cognitive dimensions. Comparing these results with zero-shot performance reveals that in-context examples affect models differently across educational tasks. most models show slightly lower average performance with few-shot prompting, challenging the assumption that examples universally improve capabilities.
Performance shifts exhibit clear task dependent patterns. Classroom dialogue classification and ethical reasoning tasks benefit from examples because they clarify task specific conventions and evaluation criteria. Conversely, creative tasks suffer as examples constrain models' generative freedom, limiting innovative thinking by anchoring responses to provided templates.
Poetry appreciation and reading comprehension show modest improvements as examples guide response formatting without necessarily enhancing comprehension. Meanwhile, memory tasks remain largely unchanged since factual recall depends primarily on pre-training knowledge, while complex reasoning often deteriorates when example reasoning paths interfere with optimal problem solving approaches.
These findings suggest educational applications should deploy few-shot prompting selectively based on specific learning objectives, recognizing that examples can either enhance or constrain model performance depending on the cognitive dimensions being evaluated.

\begin{figure}[h!]
  \centering
  \includegraphics[width=1\columnwidth, keepaspectratio]{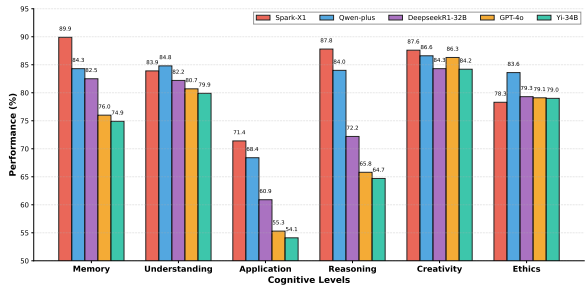} 
  \caption{The zero-shot performance of various models at different education cognitive ability levels.}
  \label{fig:zero-shot-performance}  
\end{figure}
Figure \ref{fig:zero-shot-performance} illustrates the performance of the top six models across six educational cognitive ability dimensions. From the results, we can derive several important observations.
Models demonstrate significantly stronger capabilities in Memory tasks compared to Application tasks, revealing a fundamental challenge in translating recalled knowledge into practical problem-solving scenarios. This gap suggests that while models can effectively retrieve educational content, they struggle with the dynamic application of this knowledge, particularly in classroom discourse classification, where performance drops substantially across all systems.

Performance in Reasoning and Ethics dimensions exhibits considerable variance across models. The highest performing systems achieve impressive results in logical reasoning tasks (85-90\%), but even these models show a sharp decline when confronted with multi-step reasoning problems requiring sustained logical chains. This indicates that enhancing reasoning architectures through carefully designed pretraining objectives could yield substantial improvements in educational applications requiring complex problem solving.

In the Creativity dimension, all models demonstrate moderate capabilities in content generation, yet struggle with more structured tasks like teaching design. This limitation stems primarily from the inherent tension between creative flexibility and pedagogical structure. Future development should focus on incorporating educational-specific templates and exemplars during fine-tuning to balance creative generation with instructional coherence.

Understanding dimension performance shows that models process high school level content more effectively than elementary materials, despite the latter's presumed simplicity. This counter intuitive pattern reveals a potential training distribution bias that favors more sophisticated academic content over fundamental educational concepts, suggesting that deliberate curriculum-balanced pretraining could yield more educationally versatile systems.

Ethics represents the most consistent dimension across all models, with relatively strong performance even from mid tier systems. However, this uniformity masks significant variations in handling complex ethical scenarios involving competing values or culturally sensitive educational contexts, pointing to the need for more nuanced alignment techniques that incorporate diverse educational value systems.

To address these limitations, we recommend developing specialized educational reasoning frameworks, incorporating curriculum-informed training objectives, and designing evaluation methods that better reflect authentic educational interactions. This would transform these systems from knowledge repositories into effective educational tools.

\section{Conclusion and Future Work}
We presented EduEval, a comprehensive benchmark for LLMs in Chinese K-12 education grounded in the novel EduAbility Taxonomy. Our evaluation of 14 leading models revealed a critical gap between strong factual recall and weak practical application, particularly in authentic scenarios like classroom dialogue classification. Results also showed a surprising performance hierarchy: models like Spark-X1 and Qwen-plus outperformed other leading proprietary systems in complex educational reasoning. Additionally, narrow, domain-specific fine-tuning may constrain broader cognitive abilities. These insights validate EduEval as a powerful diagnostic tool and highlight the urgent need for curriculum-balanced pre-training and improved reasoning architectures. Looking ahead, we plan to expand EduEval by adding interactive and multimodal scenarios, developing finer-grained metrics for creative tasks, and facilitating education-specific alignment strategies. We welcome community contributions to enrich this benchmark and advance AI systems that truly support diverse educational needs.
\section*{Limitations}
Despite EduEval's comprehensive assessment framework, there are several limitations. The benchmark primarily reflects Chinese educational contexts, potentially limiting international applicability. Our evaluation metrics for creativity and ethics tasks rely on GPT-based assessment, which may not fully capture the nuances of these complex cognitive dimensions despite high correlation with human judgments. The current benchmark also lacks multimodal components, overlooking the increasingly visual and interactive nature of modern education. Additionally, consistently poor performance across models in classroom dialogue classification points to significant challenges in modeling authentic teacher-student interactions. Future work will focus on incorporating multimodal tasks, developing more robust evaluation methods for higher cognitive dimensions, and expanding cross-cultural comparisons to enhance the benchmark's utility for educational applications of LLMs.
\section*{Ethics Statement}
The development of the EduEval benchmark was conducted with a strong commitment to ethical principles throughout the research process. Our ethical protocols addressed three primary areas: (1) the protection of data privacy through rigorous anonymization of all student and teacher-generated content; (2) the fair and ethical treatment of the 12 educational experts who participated in the annotation process, including fair compensation and transparent guidelines; and (3) a thorough consideration of the broader societal impacts of our work, including potential risks and advocating for the responsible use of the benchmark. A comprehensive discussion of our legal and ethical considerations is provided in Appendix \ref{appendix:Discussion}.

\bibliography{reference}

\newpage

\clearpage
\appendix

\section{Discussion}
\label{appendix:Discussion}

\subsection{Legal and Ethical Considerations.}
All data undergo strict anonymization procedures to remove personal identifiers from student content and classroom transcripts. We ensure full compliance with the General Data Protection Regulation (GDPR) and China’s Personal Information Protection Law (PIPL). The dataset is released exclusively for non-commercial academic use and has been reviewed under appropriate institutional ethics policies.
\subsection{Broader Impact}
\label{appendix:Broader-impact}

The EduEval benchmark has potential for a significant positive impact on educational technology and AI alignment with human learning processes. By providing a comprehensive evaluation framework for LLMs across diverse cognitive dimensions, our work can guide the development of more educationally relevant AI systems that better support authentic teaching and learning. The hierarchical approach of our benchmark enables targeted improvements in specific educational competencies from basic knowledge recall to complex ethical reasoning.
\subsection{Licensing Information}
However, we also acknowledge potential societal concerns. The widespread adoption of AI in educational contexts raises legitimate questions about student privacy, algorithmic bias, and the role of human teachers. There is risk that systems optimized for our benchmark could be deployed prematurely in classrooms without proper safeguards. Additionally, cultural differences in educational values and practices may not be fully represented in our Chinese education-focused framework. We encourage responsible application of our benchmark with careful consideration of these issues, and advocate for continued human oversight of AI systems in education. The EduEval benchmark should serve as a tool for enhancing human-AI collaboration in education rather than replacing human judgment in teaching and assessment.
The EduEval benchmark, including all associated datasets and source code, is made publicly available for non-commercial academic research purposes. The artifacts are released under the Creative Commons Attribution-NonCommercial-ShareAlike 4.0 International (CC BY-NC-SA 4.0) license. This license permits others to share and adapt the material for non-commercial purposes, provided that appropriate credit is given, a link to the license is provided, and any derivatives are distributed under the same license.

\section{EduAbility Taxonomy: Theoretical Foundation and Framework}
\label{appendix:eduability_framework}

This appendix summarizes the theoretical basis of the EduAbility Taxonomy and its practical use in EduEval. Traditional taxonomies offer complementary but incomplete views for educational AI assessment: Bloom’s framework emphasizes the type of cognitive operation, while DOK emphasizes the depth and rigor of tasks. In authentic classroom scenarios and model-based evaluation, cognition is multidimensional and dynamic, so a single-axis view (type-only or depth-only) is insufficient. EduAbility therefore integrates both perspectives into a compact, coherent scheme tailored for evaluating LLMs in education.

EduAbility defines six interrelated levels—Memory, Understanding, Application, Reasoning, Creativity, and Ethics—and treats them as an interconnected scaffold rather than a strict ladder. Memory and Understanding align with foundational processes and lower depth (roughly DOK 1–2, lower-order skills in Bloom). Application and Reasoning integrate strategy and multi-step inference (roughly DOK 3, Bloom’s apply/analyze/evaluate). Creativity corresponds to extended thinking and original synthesis (roughly DOK 4, Bloom’s create). Ethics is introduced as an education-specific dimension for moral judgment, fairness, privacy, and academic integrity, which is not explicitly covered by traditional cognitive taxonomies.

The scaffolded nature of the framework reflects how real tasks recruit multiple abilities at once. Higher-level tasks routinely invoke foundational knowledge; conversely, foundational tasks can be embedded in richer contexts that require interpretation or ethical judgment. This view supports precise task design and principled difficulty control while preserving authenticity. It also clarifies why performance can vary across dimensions and subjects, and how targeted improvements can be planned.

For implementation, task assignment follows a two-step principle: identify the dominant cognitive operation, then determine the required depth. Secondary processes (e.g., recalling facts within a creative task) are recorded but do not override the primary label. Ethics is applied whenever normative reasoning is integral to the correct or preferred outcome. This procedure promotes consistent classification across subjects and grade levels and facilitates reliable evaluation and analysis.

Finally, EduAbility is intended as a practical, extensible taxonomy. The current mapping and definitions are optimized for Chinese K–12 contexts and LLM assessment; as data and use cases broaden, the framework can be refined (e.g., by adding sub-levels or clarifying cross-level composites) while preserving the core integration of “operation type” and “depth” that underpins its validity and utility.

\section{Task Details}
\label{appendix:data_source}

\subsection{Data Processing Pipeline}
\label{appendix:data-processing}

We implement an agent-based pipeline that transforms 12,437 raw educational materials into 11,150 standardized benchmark items through four specialized agents. Each agent addresses a specific processing challenge while maintaining modularity and error traceability.

\subsubsection{Agent 1: Format Conversion Agent}

This agent handles heterogeneous input formats (PDFs, Word documents, LaTeX, web content) and performs intelligent OCR routing. Based on content analysis, it automatically selects: Tesseract 5.0 for standard Chinese text (98.3\% accuracy), MathPix API for mathematical notation (95.7\% accuracy), or PaddleOCR for complex layouts. Documents with OCR confidence below 85\% are flagged for human review. The agent applies post-processing rules to correct common OCR errors (e.g., 0/O, l/I/1 disambiguation). Output: 12,201 clean text representations (98.1\% retention rate).

\subsubsection{Agent 2: Standardization Agent}

This agent converts processed text into unified JSON schema containing: task ID, cognitive level, subject, grade, prompt, options, answer, explanation, difficulty, and evaluation metric. It performs three validation checks: (1) required field completeness, (2) encoding normalization to UTF-8, (3) semantic consistency using sentence-BERT to detect prompt-answer mismatches (threshold: 0.3 similarity). Items failing validation are routed to expert correction queues. This process caught 67 answer key errors. Output: 12,201 schema-compliant items (100\% pass rate).

\subsubsection{Agent 3: Task Format Optimizer}

This agent determines optimal presentation format based on cognitive requirements. For objective tasks (Memory, Understanding, Application, Reasoning, Ethics), it converts items to standardized 4-option multiple-choice format (9,250 items, 83\%). When source materials lack sufficient options, it generates distractors using rule-based heuristics derived from documented student misconceptions. For open-ended tasks (Poetry Appreciation, Reading Comprehension, Essay Scoring, Creativity), it preserves free-form structure and extracts evaluation rubrics into machine-readable format (1,900 items, 17\%). Output: 11,991 optimally formatted items.

\subsubsection{Agent 4: Quality Control Agent}

This agent implements four validation mechanisms: (1) \textbf{Duplicate detection} using MinHash LSH (85\% threshold) removes 210 redundant items; (2) \textbf{Difficulty calibration} tests items with three baseline models (GPT-4o, Qwen-72B, GLM4-9B), flagging items with $>$95\% or $<$10\% accuracy—89 outliers reviewed; (3) \textbf{Cognitive alignment verification} uses fine-tuned RoBERTa-Chinese classifier (91.3\% accuracy) to validate EduAbility level assignments, flagging 91 misclassifications; (4) \textbf{Anomaly detection} identifies statistical outliers in length, answer distribution, or content quality, catching 178 additional issues. Output: 11,150 validated items (93.0\% final retention).

\subsubsection{Agent Orchestration}

A central orchestration system coordinates agent execution, manages error handling, and routes flagged items to human reviewers. The review dashboard presents: original content, processing history, specific quality concerns, and contextual aids (e.g., scanned images for OCR errors, taxonomy definitions for cognitive misalignments). Average expert review time: 4.2 minutes per flagged item (327 items total). All agents maintain processing logs enabling full traceability from source to final item. 

\begin{table}[h]
\centering
\small
\begin{tabular}{lcc}
\toprule
\textbf{Agent} & \textbf{Input} & \textbf{Output} \\
\midrule
Format Conversion & 12,437 & 12,201 (98.1\%) \\
Standardization & 12,201 & 12,201 (100\%) \\
Format Optimizer & 12,201 & 11,991 (98.3\%) \\
Quality Control & 11,991 & 11,150 (93.0\%) \\
\bottomrule
\end{tabular}
\caption{Agent processing statistics}
\end{table}
\subsection{Task Statistics}
EdUEval includes a total of 24 distinct tasks spanning six cognitive levels as defined by the EduAbility taxonomy. As shown in Table~\ref{tab:level-stats}, each cognitive level is represented by 3–5 task types, ensuring a balanced distribution across memory, understanding, application, reasoning, creativity, and ethics. In total, the dataset comprises 11,150 instances, offering broad task diversity while maintaining a manageable data scale.

To reflect the range of linguistic and cognitive complexity present in real-world educational scenarios, Table~\ref{tab:token-statistics} provides fine-grained statistics on each task, including the number of samples and the average input length (in characters). Tasks such as \textit{Reading Comprehension} and \textit{Essay Scoring} exhibit the longest average input lengths (1001 and 819 characters, respectively), requiring models to process extended discourse. Conversely, tasks like \textit{Primary Formula Recall} and \textit{Primary Reasoning} are concise, averaging under 50 characters, and focus on factual retrieval. Creative tasks such as \textit{Text Writing} and \textit{Teaching Design} also demonstrate longer prompt structures due to their generative and open-ended nature.

This variation in task type and input complexity ensures EdUEval serves as a robust benchmark for evaluating language models under diverse educational demands and input conditions.

\begin{table}[h!]
\centering
\small
\caption{Task types and total instances per cognitive level in EdUEval.}
\label{tab:level-stats}
\begin{tabular}{lrr}
\toprule
\textbf{Cognitive Level} & \textbf{\#Task Types} & \textbf{\#Instances} \\
\midrule
Memory & 3 & 1400 \\
Understanding & 5 & 3200 \\
Application & 5 & 2600 \\
Reasoning & 4 & 2000 \\
Creativity & 3 & 200 \\
Ethics & 4 & 1750 \\
\bottomrule
\end{tabular}
\end{table}

\begin{table*}[ht]
\centering
\small
\caption{Number of instances and average input length per task in EdUEval.}
\label{tab:token-statistics}
\begin{tabular}{lllrr}
\toprule
\textbf{Level} & \textbf{Task} & \textbf{Type} & \textbf{\#Instances} & \textbf{Avg. Length (chars)} \\
\midrule
Memory & Junior Knowledge Recall & MCQ & 500 & 75 \\
Memory & Primary Formula Recall & MCQ & 400 & 43 \\
Memory & Senior Concept Recall & MCQ & 500 & 79 \\
Understanding & Junior Understanding & MCQ & 1000 & 78 \\
Understanding & Primary Understanding & MCQ & 1000 & 62 \\
Understanding & Senior Understanding & MCQ & 1000 & 167 \\
Understanding & Poetry Appreciation & MCQ & 100 & 165 \\
Understanding & Reading Comprehension & Short Answer & 100 & 1001 \\
Application & Classroom Dialogue Classification & Classification & 1000 & 62 \\
Application & Junior Problem Solving & MCQ & 500 & 71 \\
Application & Primary Problem Solving & MCQ & 500 & 38 \\
Application & Senior Problem Solving & MCQ & 500 & 74 \\
Application & Essay Scoring & Essay Score & 100 & 819 \\
Reasoning & Junior Reasoning & MCQ & 500 & 86 \\
Reasoning & General Logical Inference & MCQ & 500 & 217 \\
Reasoning & Primary Reasoning & MCQ & 500 & 49 \\
Reasoning & Senior Reasoning & MCQ & 500 & 129 \\
Creativity & Question Generation & Generation & 100 & 101 \\
Creativity & Teaching Design & Generation & 50 & 212 \\
Creativity & Text Writing & Generation & 50 & 148 \\
Ethics & Junior Ethics Scenario & MCQ & 500 & 201 \\
Ethics & Senior Ethics Scenario & MCQ & 500 & 213 \\
Ethics & Educational Ethics Judgment & MCQ & 500 & 215 \\
Ethics & Primary Moral & MCQ & 250 & 186 \\
\bottomrule
\end{tabular}
\end{table*}

\begin{table*}[h!]
\centering
\caption{Details of tasks within EduEval across six cognitive levels.}
\resizebox{\textwidth}{!}{
\label{tab:edueval-tasks}
\begin{tabular}{llp{4.8cm}llr}
\toprule
\textbf{Level} & \textbf{ID} & \textbf{Task} & \textbf{Metrics} & \textbf{Data Source} & \textbf{Test Set} \\
\midrule
Memory & 1-1 & Primary Formula Recall & Accuracy & Primary Exams & 400 \\
       & 1-2 & Junior Knowledge Recall & Accuracy & Exam Papers & 500 \\
       & 1-3 & Senior Concept Recall & Accuracy & High School Exams & 500 \\
\midrule
Understanding
              & 2-1 & Primary  Understanding & Accuracy & Primary Tests & 1000 \\
               & 2-2 & Junior  Understanding & Accuracy & Middle School Tests & 1000 \\
              & 2-3 & Senior  Understanding & Accuracy & Senior Tests & 1000 \\
              & 2-4 & Poetry Appreciation & Rouge-L & Textbook & 100 \\
              & 2-5 & Reading Comprehension & Rouge-L & Reading Bank & 100 \\
\midrule
Application & 3-1 & Classroom Dialogue Classification & Accuracy & Dialogue Corpus & 1000 \\
            & 3-2 & Primary Problem Solving & Accuracy & Primary Assignments & 500 \\
            & 3-3 & Junior  Problem Solving & Accuracy & Mock Exams & 500 \\
            & 3-4& Senior  Problem Solving & Accuracy & Primary Assignments & 500 \\
            & 3-5 & Essay Scoring & RMSE & Annotated Essays & 100 \\
\midrule
Reasoning & 4-1 & General Logical Inference & Accuracy & Logic Tasks & 500 \\
          & 4-2 & Primary  Reasoning & Accuracy & Primary Logic Tests & 500 \\
          & 4-3 & Junior  Reasoning & Accuracy & Math Puzzles & 500 \\
          & 4-4 & Senior  Reasoning & Accuracy & Senior Logic Tasks & 500 \\
\midrule
Creativity & 5-1 & Question Generation & GPT-based & Expert Annotation & 100 \\
           & 5-2 & Teaching Design & GPT-based & Expert Annotation & 50 \\
           & 5-3 & Text Writing & GPT-based & Expert Annotation & 50 \\
\midrule
Ethics & 6-1 & Primary Moral  & Accuracy & Expert Annotation & 250 \\
       & 6-2 & Junior Ethics Scenario & Accuracy & Expert Annotation & 500 \\
       & 6-3 & Senior Ethics Scenario & Accuracy & Expert Annotation & 500 \\
       & 6-4 & Educational Ethics Judgment & Accuracy & Expert Annotation & 500 \\

\bottomrule
\end{tabular}
}
\end{table*}

\subsection{Task Definition and Construction Process}
\label{appendix:task definition}
Following the EduAbility cognitive taxonomy, we designed a suite of tasks targeting six hierarchical levels of educational ability: Memory, Understanding, Application, Reasoning, Creativity, and Ethics. Each task is categorized by its dominant cognitive level, although some tasks inevitably engage multiple levels of thinking. Our classification reflects the principal skill emphasized in the prompt and expected response, ensuring consistency across task design and evaluation.

The task set includes both data-derived and expert-constructed components. Tasks aligned with Memory, Understanding, Reasoning, and Application are primarily adapted from real-world educational materials, including national and regional exams (e.g., Gaokao, Zhongkao), textbook-based assessments, and authentic classroom transcripts. In contrast, Creativity and Ethics tasks are fully designed by expert annotators to capture open-ended generation, pedagogical planning, and moral reasoning in education-specific contexts.

For tasks requiring manual creation or post-editing (e.g., classroom dialogue classification, ethical dilemmas, and all creativity tasks), we provide a high-level description of the construction workflow in this section. Annotators were trained in task-specific rubrics and followed standardized procedures to ensure validity and internal consistency. Each task is reviewed and validated by senior educators. Detailed prompt formats and example inputs are listed in Appendix~\ref{prompt-templates}, and task-specific annotation guidelines are available in Appendix~\ref {appendix:expert-annotation}.

\subsubsection{Memorization}
Tasks at the Memorization level aim to evaluate a model's ability to recall factual, curriculum-aligned knowledge across subjects and educational stages. These tasks emphasize short-context retrieval and are presented in multiple-choice format. The design spans primary to high school content and reflects core requirements from national educational standards. This section includes the following three tasks:
 \paragraph{Primary Formula Recall (1-1)}  
This task focuses on foundational knowledge from primary education, especially in Mathematics and the Chinese Language. Questions include recalling area and volume formulas, converting measurement units, and recognizing character pronunciations, pinyin, common radicals, and mythological references. A total of 400 items were selected from standardized primary-level assessments to ensure alignment with the textbook scope and learning expectations.

\paragraph{Junior Knowledge Recall (1-2)}  
This task includes factual items from middle school subjects such as Science, Geography, History, and Information Technology. Examples include identifying the function of a keyboard shortcut, naming a historical figure, or recalling a chemical definition. The content emphasizes breadth across disciplines and regional test styles. A total of 500 questions were collected from official mock exams and public test banks.

\paragraph{Senior Concept Recall (1-3)}  
This task targets advanced memorization needs in high school. Disciplines include Physics (e.g., electromagnetic laws), Chemistry (e.g., organic reaction equations), Mathematics (e.g., trigonometric identities), Chinese (e.g., classical prose interpretation), and English (e.g., complex sentence structures and grammar patterns). All questions are extracted from real high-stakes exams such as Gaokao mock papers. The final dataset contains 500 questions with subject-balanced coverage.

These tasks emphasize accurate reproduction of textbook-based content and offer a foundation for evaluating whether language models can serve as reliable aids for early-stage educational reinforcement.

\subsubsection{Understanding}
Understanding-level tasks aim to assess a model’s ability to process and internalize meaning from structured educational texts. These tasks require the model to grasp explicit and implicit information, interpret rhetorical or scientific intent, and identify relationships between concepts or events. Unlike simple recall, understanding involves cognitive operations such as summarizing main ideas, identifying supporting details, recognizing cause-and-effect structures, interpreting literary techniques, and distinguishing between factual statements and inferred meanings.
The task design spans diverse genres—narrative, expository, poetic, and scientific—across different grade levels. Primary-level tasks focus on literal comprehension and simple reasoning; middle-school tasks introduce structural logic and semantic nuance; high-school tasks emphasize conceptual depth and domain-specific textual analysis. Formats include both multiple-choice and short-answer questions, depending on the task objective and response requirements.
\paragraph{Primary Understanding (2-1)}  
At the elementary level, understanding is centered on grasping literal meaning, everyday scenarios, and simple inferencing. Selected texts use clear sentence structures, familiar vocabulary, and relatable themes such as family, school, or animals. Questions require identification of key facts, character actions, or intentions. All content is aligned with primary school language curricula and sourced from official exam papers and children's reading materials.
\paragraph{Junior Understanding (2-2)}  
Designed for the middle school level, this task assesses a model’s ability to identify central ideas, extract factual details, and recognize logical relationships in expository and narrative texts. Question design emphasizes structural comprehension, temporal or causal coherence, and contrastive reasoning. Texts are drawn from Chinese language textbooks, educational supplements, and online learning platforms to ensure age-appropriate difficulty and domain coverage.

\paragraph{Senior Understanding (2-3)}  
 Comprehension at the high school level involves interpreting disciplinary texts across subjects such as Physics, Chemistry, Biology, History, and Political Science. Models must decode technical definitions, reason through scientific arguments, and synthesize multi-paragraph information. Texts are adapted from national curricula, Gaokao mock exams, and subject-specific academic articles to reflect cognitively demanding reading expected in upper secondary education.
\paragraph{Poetry Appreciation (2-4)}  
Poetic comprehension tasks evaluate the ability to interpret imagery, tone, structure, and rhetorical devices within classical or modern Chinese poetry. Questions probe a model’s grasp of metaphor, thematic progression, and stylistic nuance across different eras and genres. All poems are selected from high school literature textbooks and curated poetry collections. Responses are formulated in short-answer format to reflect interpretive depth.
\paragraph{Reading Comprehension (2-5)}  
Long-form reading comprehension tasks assess multi-layered understanding of essays, nonfiction articles, and other extended texts. Questions are designed to probe core propositions, supporting arguments, logical structure, and implicit messages. Source texts are selected from authentic reading materials used in educational contexts. Responses are open-ended, and model output is evaluated with ROUGE-L to assess semantic coverage and precision.

These tasks are essential for evaluating how well language models can replicate the comprehension skills taught in K–12 curricula and adapt to varying difficulty levels across reading and subject domains.

\subsubsection{Application}
Application-level tasks evaluate a model’s ability to transfer learned knowledge to practical, context-rich educational settings. These tasks simulate authentic classroom scenarios where students are expected to solve problems, interpret structured information, make pedagogical classifications, or assess open-ended work. The task set spans multimodal dialogue interpretation, quantitative reasoning, and evaluative writing feedback across primary to high school levels.

Classroom classification tasks require models to categorize utterances into instructional roles and are evaluated using standard classification accuracy. Essay scoring tasks demand nuanced judgments along multiple dimensions such as coherence, grammar, and content relevance; models are expected to assign scalar scores aligned with expert-annotated rubrics. Mathematical problem-solving tasks, ranging from basic arithmetic to multi-step scientific applications, assess models’ ability to process numerical and symbolic input in realistic educational formats. Depending on the task structure, responses are evaluated via accuracy for closed-form answers or correlation-based metrics (e.g., QWK, Pearson) for open-ended assessment.

\paragraph{Classroom Dialogue Classification (3-1)}  
This task involves analyzing authentic classroom dialogue transcripts and classifying utterances into pedagogically meaningful categories (e.g., teacher explanation, student elaboration, questioning, or feedback). The dialogues are sourced from real middle school classrooms in subjects such as Information Technology, Mathematics, and Chinese. Original classroom recordings were transcribed to text and annotated by domain experts. The detailed taxonomy used for classification is provided in Appendix~\ref{appendix:expert-annotation}.
 \paragraph{Primary Problem Solving (3-2)}  
The primary-level application tasks focus on basic quantitative reasoning grounded in everyday life. Problem types include simple arithmetic word problems, time-distance calculations, shopping-related questions, and basic engineering/logistics scenarios adapted to the cognitive level of elementary students. Content is curated from elementary school test banks and real classroom assessments.
\paragraph{Junior Problem Solving (3-3)}  
Tasks at the junior high level include typical application-oriented problems in Mathematics, Physics, and Chemistry. These questions involve numerical calculation, data interpretation, and conceptual application in scenarios such as physical motion, chemical change, and geometric modeling. All questions are sourced from historical Zhongkao exams, regional mock tests, and online education platforms, reflecting the procedural rigor of national curriculum standards.

\paragraph{Senior Problem Solving (3-4)}  
At the high school level, application tasks require multi-step reasoning, cross-topic integration, and realistic problem contexts. Problems often span several knowledge points across disciplines—for example, combining algebra and geometry, or interpreting physics data within real-life scientific experiments. Questions are selected from Gaokao-level assessments and authentic academic preparation resources, representing the most complex application scenarios in the dataset.

\paragraph{Essay Scoring (3-5)}  
This task evaluates the model’s ability to assign scores to student-written essays in accordance with expert grading. Essays are collected from real classroom writing assignments in primary and middle school settings, covering narrative, argumentative, and descriptive genres. Each essay is paired with a reference score given by certified instructors following standard rubrics. Model responses are evaluated using QWK and Pearson correlation with human scores.

\subsubsection{Reasoning}
Reasoning-level tasks evaluate a model’s capacity for deductive, multi-step, and abstract thinking across various educational contexts. These tasks require more than fact recall or surface comprehension—they challenge the model to infer, calculate, compare, or generalize based on limited or structured input. Reasoning items span all school stages and are primarily designed as multiple-choice questions. While some are discipline-specific, others are purely logical in nature, simulating standardized test conditions.
\paragraph{General Logical Inference (4-1)}  
Independent of school subjects, this task focuses purely on formal logic and critical reasoning. Question types include identifying assumptions, strengthening/weakening arguments, and detecting logical fallacies. For example, a question may require challenging a conclusion about rainforest preservation based on environmental data or identifying which option undermines a policy argument. The format follows logical reasoning tests common in aptitude exams, using real-world scenarios from domains such as economics, environmental policy, and public health. Difficulty is moderate to slightly above moderate, with items often requiring multi-constraint deduction.
\paragraph{Primary Reasoning (4-2)}  
Designed for elementary-level learners, this task emphasizes foundational reasoning within Mathematics and Chinese. Questions involve logical enumeration, basic inference, and intuitive problem solving. For instance, items include "What is the smallest number divisible by 9 that leaves a remainder of 5 when divided by 6?", or "How many handshakes occur between three friends?". Other questions explore creative problems such as color-painting cubes or firecracker arrangements. Most items are labeled as "easy", with a few rising to "moderate", and maintain age-appropriate vocabulary and structure.
\paragraph{Junior Reasoning (4-3)}  
This task targets middle school students’ reasoning ability across nine disciplines: Chinese, Mathematics, English, Physics, Chemistry, Biology, History, Geography, and Civics. Items assess symbolic manipulation, spatial logic, experimental deduction, and semantic judgment. For example, Chinese questions involve sentence pair matching and rhetorical inference; geography questions test spatial reasoning, such as identifying provinces on a map. Chemistry problems may involve reactivity series, while physics questions assess optical phenomena (e.g., convex lens imaging). The overall difficulty is labeled as "easy to moderate," reflecting the cognitive profile of middle school learners.

\paragraph{Senior Reasoning (4-4)}  
This task presents high school students with complex, multi-disciplinary reasoning challenges. It includes domain-specific inference in subjects such as Mathematics (e.g., complex numbers, probability, normal distribution), Physics (e.g., electromagnetic modeling), Biology (e.g., genetics and heredity), and Political Science (e.g., interest rate policy chains). For example, a chemistry question may involve interpreting electrochemical recovery of aluminum alloys; a history question may ask for reasoning behind defining the "14-year War of Resistance". These questions integrate real-world applications and often require synthesis of multiple knowledge points, with difficulty ranging from "moderate" to "difficult".

These tasks test a language model’s ability to handle structured constraints, domain-specific reasoning, and logical abstraction. They simulate classroom and exam conditions where inference, rather than memory, is the central demand.
\subsubsection{Creativity}
Creativity-level tasks assess the model’s capacity to generate novel, coherent, and contextually appropriate content across a range of educational domains. These tasks move beyond recall or selection and require open-ended, pedagogically aligned language generation. All prompts are manually constructed by experienced educators to ensure diversity, authenticity, and alignment with modern instructional goals. Responses are evaluated using GPT-based scoring models to assess creativity, semantic relevance, and pedagogical appropriateness.

\paragraph{Question Generation (5-1)}  
This task involves generating original, subject-specific questions for students from primary to high school. The prompts emphasize cross-disciplinary design, encouraging the model to embed real-world contexts such as shopping scenarios in math or historical debates in social studies. The task targets the model’s ability to foster critical thinking, practical reasoning, and curriculum-oriented assessment skills. Question quality is evaluated by large language models along dimensions such as clarity, cognitive demand, and alignment with instructional objectives.

\paragraph{Teaching Design (5-2)}  
 Given a specific subject, topic, and grade level, the model is required to propose a detailed teaching plan. Each plan should include learning objectives, instructional strategies, classroom activities, and assessment methods. The prompt structure reflects best practices in instructional design and encourages active, student-centered pedagogy. This task evaluates the model’s instructional planning ability under realistic teaching constraints. Responses are automatically rated by GPT-based evaluators for pedagogical soundness, structure, and feasibility.

\paragraph{Text Writing (5-3)}  
 Writing tasks span a wide variety of genre,s including narrative essays, poetic compositions, argumentative texts, and technical explanations. Prompts are designed to cover different proficiency levels and stylistic requirements. The model must demonstrate creativity, genre awareness, and communicative clarity. Each response is assessed using large model-based metrics that measure semantic richness, structural coherence, and stylistic appropriateness across genres.

By leveraging GPT-based evaluation models, we ensure a scalable, consistent assessment of creative outputs across open-ended tasks, covering multiple educational levels and writing purposes.

\subsubsection{Ethics}
Ethics-level tasks are designed to assess a model’s ability to make morally informed judgments and reflect normative reasoning in educational scenarios. The tasks simulate real-world ethical dilemmas or value-based decision-making processes, covering individual conduct, social behavior, and professional responsibility. Prompts are carefully constructed by experts in moral education and educational psychology, with responses evaluated based on alignment with educational norms, fairness, and contextual sensitivity.

Ethics-level tasks are designed to assess a model’s ability to make morally informed judgments and reflect normative reasoning in educational scenarios. These tasks simulate real-world ethical dilemmas involving individual behavior, community responsibility, and professional values. Each task is carefully constructed by educational or ethics experts and evaluated based on alignment with commonly accepted moral standards in school or societal contexts.

\paragraph{Primary Moral (6-1)}  
This task focuses on elementary-level moral understanding. Scenarios involve basic ethical choices such as honesty, cooperation, empathy, and rule-following. The goal is to evaluate whether a model can correctly distinguish right from wrong in age-appropriate situations, including classroom behavior or helping others in community settings. All questions are multiple-choice and presented in simplified language suitable for primary students.

\paragraph{Junior Ethics Scenario (6-2)}  
At the junior high level, ethical dilemmas are framed around interpersonal and family issues, such as fairness, peer conflict, and communication. Questions are designed to reflect the psychosocial development stage of adolescents and assess their reasoning about relationships, social roles, and responsibility in group contexts.

\paragraph{Senior Ethics Scenario (6-3)}  
Senior high school tasks expand into themes of social responsibility, civic engagement, and ethical leadership. Scenarios require reasoning about digital behavior, public policy, or societal norms—for example, evaluating the ethics of reposting sensitive content online, or considering actions to mitigate climate change. These tasks demand higher-order moral judgment and trade-off evaluation.

\paragraph{Educational Ethics Judgment (6-4)}  
This task simulates ethical decision-making in educational professional settings, such as classroom equity, grading fairness, and student-teacher dynamics. It is intended for evaluating adult or pre-service teacher alignment with ethical codes in educational practice. Questions include dilemmas involving bias, academic honesty, and responsibility toward students’ well-being.

\subsection{Task Instruction and Example}
\label{prompt-templates}
In this section, we provide representative input-output examples for each task category in EdUEval. To ensure scalability and consistency across diverse tasks, we adopt a unified prompt format where possible, indicating the task description, input query, and expected response structure.
Examples are grouped by task type and cognitive level. Tables~\ref{tab:example-senior-mcq} illustrate multiple-choice tasks across various domains, covering memorization, understanding, application, and reasoning. Poetry appreciation tasks are shown in Table~\ref{tab:example-poetry}, while Table~\ref{tab:example-reading} presents the prompt and annotated answer format used for reading comprehension tasks.

We further include examples of complex, real-world educational tasks. Table~\ref{tab:example-dialogue} illustrates classroom dialogue classification based on authentic teacher-student transcripts. Table~\ref{tab:example-essay} demonstrates how student-written compositions are paired with rubrics and model scoring. For open-ended generation tasks, Table~\ref{tab:example-question} displays question generation prompts, Table~\ref{tab:example-teaching} outlines a teaching design example, and Table~\ref{tab:example-writing} provides a complete instance of model-generated text writing.

These examples serve as reference points for users and researchers to better understand the structure and intent of each task format in EdUEval.

\subsection{Comparison with Existing Benchmarks}

\begin{table*}[htbp]
\centering
\caption{Comparison of EdUEval with existing educational benchmarks.}
\label{tab:benchmark_comparison}
\begin{tabular}{lccccc}
\toprule
\textbf{Benchmark} & \textbf{Chinese} & \textbf{Cognitive} & \textbf{Authentic} & \textbf{Creative/} & \textbf{Scale} \\
 & \textbf{Coverage} & \textbf{Levels} & \textbf{Tasks} & \textbf{Ethical} & \textbf{(\# Questions)} \\
\midrule
MMLU & Partial & 2--3 & No & No & 15,908 \\
BIG-bench & Partial & 2--4 & No & Limited & 204,000+ \\
HELM & Partial & 2--3 & Limited & No & 16,400+ \\
MMCU & Full & 2--3 & No & No & 11,528 \\
C-Eval & Full & 2--3 & Limited & No & 13,948 \\
CMMLU & Full & 2--3 & No & No & 11,528 \\
AGIEval & Full & 2--4 & No & No & 8,062 \\
E-Eval & Full & 2--4 & Limited & No & 18,717 \\
\textbf{EdUEval (ours)} & \textbf{Full} & \textbf{1--6} & \textbf{Yes} & \textbf{Yes} & \textbf{11,150} \\
\bottomrule
\end{tabular}
\end{table*}

\subsection{GPT-based Evaluation Protocols}
\label{appendix:gpt-eval}
To evaluate open-ended tasks such as text writing, question generation, and teaching design, we adopt GPT-based scoring as a consistent and scalable alternative to human annotation. We prompt GPT-4 with task-specific rubrics and set the decoding temperature to zero to ensure deterministic output. Each rubric contains clearly defined criteria across multiple dimensions, such as content accuracy, structural coherence, pedagogical relevance, creativity, and feasibility. The model receives both the input content and the full rubric and returns a holistic score ranging from 0 to 100 without explanation. The detailed scoring rubrics used for each task type are shown in Tables~\ref{tab:gpt-teaching-rubric} to~\ref{tab:gpt-writing-rubric}. To validate reliability, we sample 100 examples and compare GPT-4 scores with human expert annotations, observing Pearson correlations above 0.87 across all tasks. This automated evaluation framework allows efficient and robust benchmarking of model-generated outputs in educational settings while maintaining alignment with human grading standards.

\section{Experimental Details}
\label{appendix:experimental-setup}

All evaluation experiments were conducted on a high-performance computing server equipped with 3 NVIDIA GPUs (2 × GeForce RTX 3090 with 24576MB VRAM each and 1 × RTX A6000 with 49140MB VRAM) plus 4 additional NVIDIA GeForce RTX 4090 GPUs, all powered by a 128-core Intel(R) Xeon(R) Platinum 8358 CPU @ 2.60GHz. Our software environment consisted of Ubuntu 22.04 LTS with CUDA 12.7, Python 3.9.0, PyTorch 2.0.0, and Transformers 4.28.1. We utilized the RTX A6000 primarily for resource-intensive few-shot evaluations due to its larger VRAM capacity, while distributing smaller model evaluations and preprocessing tasks across the RTX 3090 and RTX 4090 GPUs to ensure consistent and fair performance comparison across all cognitive dimensions.

\begin{table}[ht]
\renewcommand{\arraystretch}{1.2}
\centering
\small
\caption{Models evaluated in this paper.}
\begin{tabularx}{\columnwidth}{lXcc}
\toprule
\textbf{Model} & \textbf{\#Parameters} & \textbf{Access} \\
\midrule
GPT-4o  & undisclosed & API \\
Spark-X1  & undisclosed & API \\
Qwen-plus  & undisclosed & API \\
Qwen-72B  & 72B & Weights \\
Qwen-14B-Chat  & 14B & Weights \\
Qwen-7B & 7B & Weights \\
Yi-34B-Chat  & 34B & Weights \\
Yi-6B-Chat  & 6B & Weights \\
Chinese-LLaMA2-13B  & 13B & Weights \\
LLaMA3-8B  & 8B & Weights \\
GLM4-9B-Chat  & 9B & Weights \\
DeepSeekR1-32B  & 32B & Weights \\
EduChat-sft-002-13B & 13B & Weights \\
BaiChuan-13B & 13B & Weights \\
\bottomrule
\end{tabularx}
\label{tab:evaluated-models}
\end{table}

\section{Guidelines for Expert Annotation}
\label{appendix:expert-annotation}

The expert annotation protocol was designed to systematically operationalize the EduAbility cognitive scaffold through rigorous human judgment, ensuring that each benchmark item is not only pedagogically sound but also theoretically grounded. To this end, we recruited 12 educational experts with an average of 8.5 years of teaching experience across primary, secondary, and higher education. All annotators possessed formal pedagogical training and curriculum development experience within the Chinese educational system, making them uniquely qualified to interpret and apply our nuanced cognitive framework.

The core of the annotation process was a multi-stage validation pipeline, beginning with intensive training where experts were familiarized with the theoretical underpinnings of the EduAbility Taxonomy. The primary directive for all tasks was to identify and label the \textbf{dominant cognitive operation} required by an item, thereby mapping it to a specific level of the cognitive scaffold.

Task-specific guidelines were established to ensure this principle was applied consistently:
\textbf{For Classroom Dialogue Classification,} annotators were instructed to move beyond surface-level semantics and analyze the full conversational flow---including discourse function, speaker intention, and pedagogical role. This holistic analysis was essential for classifying utterances according to their function within the 
\textbf{Application} level of our framework (e.g., Guidance, Inference vs. simple Knowledge Recall). Ambiguous utterances were flagged for consensus discussions to maintain theoretical alignment.
\textbf{For Essay Scoring tasks,} experts rated compositions using official grading rubrics that serve as practical proxies for the \textbf{Application} and \textbf{Creativity} dimensions of the taxonomy. Dimensions such as content, coherence, and expressiveness were evaluated to produce a holistic score. To ensure inter-rater reliability, annotators calibrated their scoring in periodic review meetings, and only essays with consensus scores or a variance within \(\pm2\) points were retained.
\textbf{For Ethical Judgment tasks,} the annotation process directly reflected the design of the \textbf{Ethics} dimension in our taxonomy, which prioritizes nuanced moral reasoning over simple right/wrong dichotomies. Consequently, these tasks were not labeled with "correct" answers. Instead, experts constructed the full scenario and option set to be pedagogically plausible and reflect genuine moral ambiguity, drawing from real-world dilemmas in teaching and school policy.

To ensure the integrity of this process, participants were compensated at a rate of ¥150 (about \$21) per hour and completed a paid training session before formal annotation began. Across all tasks, consistency was maintained through standardized glossaries and shared exemplars. All annotations underwent multi-round review by senior experts, with correction logs kept to trace decision rationale. This rigorous, human-led process was the crucial preparatory step for the final AI-driven validation performed by the \textbf{Cognitive Alignment Verification Agent} (as described in Section 3.2), ensuring that the final dataset is a faithful operationalization of the EduAbility framework.







\begin{table}
\centering\small
\begin{tabular}{p{0.65\linewidth} r}
\toprule
\textbf{Scoring Dimension} & \textbf{Points} \\
\midrule
\textbf{Accuracy and Clarity} & 20 \\
\quad Content alignment with subject knowledge & 10 \\
\quad Clear and unambiguous phrasing & 10 \\
\cmidrule{1-2}
\textbf{Educational Value} & 20 \\
\quad Core concept coverage & 10 \\
\quad Cognitive demand (e.g., Bloom's levels) & 10 \\
\cmidrule{1-2}
\textbf{Task Design Quality} & 30 \\
\quad Appropriate question format & 10 \\
\quad Realistic and engaging context & 10 \\
\quad Depth and integrative reasoning & 10 \\
\cmidrule{1-2}
\textbf{Difficulty and Challenge} & 20 \\
\quad Proper difficulty gradient & 10 \\
\quad Opportunities for divergent thinking & 10 \\
\cmidrule{1-2}
\textbf{Originality and Utility} & 10 \\
\quad Novel framing or perspective & 5 \\
\quad Practical instructional usefulness & 5 \\
\midrule
\textbf{Total} & \textbf{100} \\
\bottomrule
\end{tabular}
\caption{GPT-based scoring rubric for Question Generation tasks.}
\label{tab:gpt-qg-rubric}
\end{table}

\begin{table}
\centering\small
\begin{tabular}{p{0.65\linewidth} r}
\toprule
\textbf{Scoring Dimension} & \textbf{Points} \\
\midrule
\textbf{Language Accuracy} & 20 \\
\quad Grammar correctness & 10 \\
\quad Vocabulary precision & 10 \\
\cmidrule{1-2}
\textbf{Content Quality} & 30 \\
\quad Richness and specificity & 10 \\
\quad Coherence and flow & 10 \\
\quad Thematic depth & 10 \\
\cmidrule{1-2}
\textbf{Creativity and Thinking} & 30 \\
\quad Originality and novel expression & 10 \\
\quad Critical thinking and perspective & 10 \\
\quad Emotional resonance & 10 \\
\cmidrule{1-2}
\textbf{Structure and Format} & 20 \\
\quad Complete essay structure & 10 \\
\quad Formatting and conventions & 10 \\
\midrule
\textbf{Total} & \textbf{100} \\
\bottomrule
\end{tabular}
\caption{GPT-based scoring rubric for Text Writing tasks.}
\label{tab:gpt-writing-rubric}
\end{table}

\begin{table*}
\centering
\small
\begin{tabular}{p{0.95\linewidth}}
\toprule
\textbf{INSTRUCTION:} Read the following classical Chinese poem and answer the short-answer questions in concise and coherent language. \\
\midrule
\textbf{QUERY:} \textit{Poem:}  
\\
\quad \textbf{On the Eastern Tower of Xianyang} \\
\quad \textit{by Xu Hun (Tang Dynasty)} \\
\quad Climbing high brings endless sorrow, \\
\quad Willows and reeds float like isles in water. \\
\quad Clouds rise by the creek, the sun falls behind the pavilion, \\
\quad Storm winds fill the tower—rain approaches the mountain. \\
\quad Birds soar over green fields at dusk near Qin gardens, \\
\quad Cicadas cry in yellow leaves around the Han palace. \\
\quad Ask not of the past— \\
\quad The Weishui still flows east through my ruined homeland. \\
\\
(1) What images are used in the third couplet to convey desolation and bleakness? \\
(2) How does the fourth couplet express the poet’s reflection on historical change? \\
\midrule
\textbf{ANSWER:} \\
(1) Cloud, setting sun, mountain rain, and wind. \\
(2) The poet uses imagery of “birds flying over green fields” and “cicadas crying in yellow leaves” to depict the desolate atmosphere of autumn. By referencing the ruined states of the Qin and Han dynasties, he conveys deep sorrow over the passage of time and historical decline. \\
\bottomrule
\end{tabular}
\caption{Example from Poetry Appreciation task.}
\label{tab:example-poetry}
\end{table*}

\begin{table*}
\centering
\small
\begin{tabular}{p{0.95\linewidth}}
\toprule
\textbf{INSTRUCTION:} Please answer the following multiple-choice question. Provide only the letter (e.g., A, B, C, or D). \\
\midrule
\textbf{QUERY:} Which of the following is the core city of the Northeastern Atlantic coastal urban cluster in the United States? \\
A. Chicago \\
B. New York \\
C. Washington \\
D. Boston \\
\midrule
\textbf{ANSWER:} B \\
\bottomrule
\end{tabular}
\caption{Example from Senior Concept Recall task.}
\label{tab:example-senior-mcq}
\end{table*}

\begin{table*}
\centering
\small
\begin{tabular}{p{0.95\linewidth}}
\toprule
\textbf{INSTRUCTION:} Read the passage carefully and answer the following short-answer questions. Each response should be concise and directly based on the passage content. \\
\midrule
\textbf{PASSAGE (excerpt):}  
\textit{“When we arrived at Gubei Water Town, the glow of sunset had faded and the town began to light up. The night enveloped the town in deep shadow. Gubei derives its name from the Gubeikou Great Wall, composed of famous sections like Wohu Mountain and Simatai, steeped in history and heroic imagery. Once a military stronghold linking the Liaodong Plain and Inner Mongolia, it is now peaceful and serene. Despite being in northern China, it resembles a southern water town with flowing rivers, weeping willows, and spring blossoms. From Simatai, lantern-lit Great Wall silhouettes glow as the town sparkles below. Returning to town, lively street cries, child laughter, and wood-structured buildings evoke an ancient charm, while the town remains modern with creative shops and music echoing into the night.”} \\

\textbf{QUESTIONS:} \\
(1) What are the main characteristics of Gubei Water Town described in the passage? \\
(2) Analyze the language used to portray the liveliness of the town at night. \\
(3) What is the function of the inserted historical background? \\
(4) Why is the town described as a “reversed relocation of regional geography” ? \\

\midrule
\textbf{ANSWER:} \\
(1) Peaceful, elegant, lively, with Jiangnan-style charm. \\
(2) The bustling market sounds and interactions create a vibrant and immersive atmosphere. \\
(3) The historical context enriches the cultural depth and contrast between past and present. \\
(4) Though in the north, it replicates the southern water town scenery, hence the metaphor. \\

\bottomrule
\end{tabular}
\caption{Example from Reading Comprehension task.}
\label{tab:example-reading}
\end{table*}

\begin{table*}
\centering
\small
\begin{tabular}{p{0.95\linewidth}}
\toprule
\textbf{INSTRUCTION:}  
Given the classroom dialogue content below, please determine its pedagogical category. Return only the corresponding category number (1–9). Categories include: \\
1. Basic Knowledge \quad 2. Personal Information \quad 3. Analysis \quad 4. Generalization \quad 5. Inference and Transfer \\
6. Extension and Response \quad 7. Agreement \quad 8. Doubt \quad 9. Guidance \\
\midrule
\textbf{DIALOGUE:} \\
"Well, numbers have addition and subtraction, right? Do vectors also have addition and subtraction? Please reflect based on your prior knowledge of displacement and force composition, and explore this at home. Lastly, I wish you all the best in life. No matter how winding the road ahead, as long as the direction is right, go forward with all your heart and share your happiness. Class dismissed." \\
\midrule
\textbf{ANSWER:} 9 \\
\bottomrule
\end{tabular}
\caption{Example from Classroom Dialogue Classification task.}
\label{tab:example-dialogue}
\end{table*}

\begin{table*}
\centering
\small
\begin{tabular}{p{0.95\linewidth}}
\toprule
\textbf{INSTRUCTION:} Please assign a holistic score (0–100) to the following student essay. Do not provide any explanation. \\
\midrule
\textbf{ESSAY PROMPT:} \textit{The Calendar of Life} \\
\textbf{STUDENT RESPONSE:} \\
\textit{The Calendar of Life} \\
\textit{Tearing off one page at a time, each one carrying your hope.} \\
—I don't remember how long it has been since I last visited her. Childhood memories are fading, but I still recall her silver hair and the quiet traces of time etched into her face. \\
One morning, a ringing phone woke me. As I hung up, I froze—Grandma had been in a car accident. When I rushed to the hospital, she was already in the ICU. Tears flooded my eyes. Following Grandpa’s instructions, I returned to our old home to retrieve some clothes. \\
The house was the same one filled with my childhood memories. Hesitantly, I stepped up to the attic and into her room. The scent—her scent—rushed at me. On the desk, I saw large wall calendars torn apart, the pages scattered. On the back of each page were handwritten notes: “Feb 2003, Nannan came home today.” “May 2003, Nannan called and told me to dress warmly.” “July 2004, Nannan’s entrance exam.” “Feb 2006, it’s been a long time since Nannan called me.” \\
I frantically turned each page. Each recorded my visits, my calls, our moments. I imagined her, night after night, under the dim lamp, her eyes behind reading glasses, noting down everything with love. Had I grown too busy, too distant? And yet she clung to hope through these calendar pages, waiting for the little girl who used to call her “Nannan.” \\
I gathered the pages and returned to the hospital. There she lay—bandaged, asleep—yet still the face that once spoiled me. I took one last page, filled with all my longing, and placed it in her hand. “Grandma,” I whispered, “please wake up. Nannan’s here now. You don’t need to count the days anymore...” \\
\midrule
\textbf{ANSWER:} 90 \\
\bottomrule
\end{tabular}
\caption{Example from Essay Scoring task.}
\label{tab:example-essay}
\end{table*}

\begin{table*}
\centering
\small
\begin{tabular}{p{0.95\linewidth}}
\toprule
\textbf{INSTRUCTION:} Please design an educational assessment question based on the following requirements. Ensure the question is clear, curriculum-aligned, and suitable for the specified grade level. \\
\midrule
\textbf{INPUT:} \\
\textbf{Grade:} 5 \\
\textbf{Subject:} Geography \\
\textbf{Knowledge Point:} Overview of World Geography \\
\textbf{Task Description:} Create a fill-in-the-blank question on world geography. It should cover major countries on each continent, label locations of well-known mountains and rivers, and include causal reasoning about geographical phenomena. \\
\midrule
\textbf{OUTPUT:} \\
Fill in the blanks below based on your knowledge of world geography: \\
(1) The \underline{\hspace{3cm}} River is the longest river in Africa. \\
(2) The \underline{\hspace{3cm}} Mountains separate Europe from Asia. \\
(3) Brazil is located on the continent of \underline{\hspace{3cm}}. \\
(4) Earthquakes are frequent along the Pacific Ring of Fire because of \underline{\hspace{5cm}}. \\
\bottomrule
\end{tabular}
\caption{Example from Question Generation task.}
\label{tab:example-question}
\end{table*}

\begin{table*}
\centering
\small
\begin{tabular}{p{0.95\linewidth}}
\toprule
\textbf{INSTRUCTION:} Please create a detailed teaching plan based on the provided grade level, topic, and instructional goals. The design should include objectives, teaching methods, procedural steps, and resource requirements. \\
\midrule
\textbf{INPUT:} \\
\textbf{Grade:} 2 \\
\textbf{Subject:} English \\
\textbf{Topic:} Basic Grammar and Sentence Patterns \\
\textbf{Teaching Requirements:} Design a lesson to teach the basic use of the simple present tense. Students should learn to correctly use simple sentence structures. Adopt discovery-based and practice-based approaches. Teachers should provide example sentences and guide students to find grammar rules. Students will complete substitution and creative exercises. The teaching flow includes: \\
• \textit{Introduction:} Present lively scenes using simple English sentences. \\
• \textit{Example Analysis:} Show example sentences using the simple present. \\
• \textit{Rule Discovery:} Guide students to extract grammatical patterns. \\
• \textit{Substitution Practice:} Replace subjects or verbs using a worksheet. \\
• \textit{Creative Practice:} Compose short texts using the target grammar. \\
• \textit{Summary and Feedback:} Review student work and reinforce key points. \\
\textbf{Resources:} Grammar worksheets. \\
\midrule
\textbf{OUTPUT:} \\
\textbf{Lesson Title:} Exploring the Simple Present Tense \\
\textbf{Grade Level:} 2 \quad \textbf{Duration:} 40 minutes \\
\textbf{Objectives:} \\
• Understand the basic use of the simple present tense. \\
• Identify and apply sentence patterns with regular subjects and verbs. \\
• Develop the ability to modify and create simple sentences. \\

\textbf{Teaching Procedure:} \\
1. \textbf{Warm-up (5 min):} Present engaging pictures (e.g., a dog runs, a boy eats an apple) and read sentences aloud. Ask students to guess the meaning. \\
2. \textbf{Example Analysis (5 min):} Write examples on the board: "He plays soccer." "They eat lunch." Highlight subject–verb patterns. \\
3. \textbf{Grammar Discovery (5 min):} Guide students to summarize that verbs change with the subject in present tense (e.g., +s/es). \\
4. \textbf{Substitution Practice (10 min):} Use worksheets to replace subjects or verbs in given sentences. Peer check answers. \\
5. \textbf{Creative Practice (10 min):} Students compose 3–5 sentence mini-paragraphs using the learned grammar. \\
6. \textbf{Review (5 min):} Volunteers share their writing. Teacher gives positive feedback and recaps the grammar rule. \\
\textbf{Materials:} Illustrated sentence cards, grammar worksheet. \\
\bottomrule
\end{tabular}
\caption{Example from Teaching Design task.}
\label{tab:example-teaching}
\end{table*}

\begin{table*}
\centering
\small
\begin{tabular}{p{0.95\linewidth}}
\toprule
\textbf{INSTRUCTION:} Based on the following writing prompt, compose a complete short story of approximately 800 characters. The writing should be coherent, creative, and logically consistent with the given opening. \\
\midrule
\textbf{PROMPT:} \\
You are given the beginning of a story. Please continue it into a complete narrative. \\
\textit{Opening:} \\
Once there was a little boy named Pinpin. He was often bullied for being clumsy, until one day... \\
\midrule
\textbf{OUTPUT:} \\
...he found an old box behind his school, glowing faintly under the setting sun. Inside lay a strange pair of glasses. When Pinpin put them on, the world changed—he could suddenly see patterns in how things moved, how people talked, and even how puzzles worked. At first he was scared, but soon he realized this was his gift. \\
\\
The next day at school, during math class, the teacher presented a tough logic problem. Pinpin raised his hand, something he’d never done before. He calmly walked to the board and solved it, step by step, as the whole class watched in awe. The bullies were speechless. \\
\\
As days passed, Pinpin grew more confident. He used the glasses not to show off, but to help others—he taught his friends how to understand math, comforted classmates when they felt stuck, and even wrote poems that made people smile. \\
\\
One day, he returned to the place where he found the glasses. They were gone. But Pinpin smiled. He didn’t need them anymore. What he discovered wasn’t magic in the glasses—it was magic in himself. \\
\bottomrule
\end{tabular}
\caption{Example from Text Writing task.}
\label{tab:example-writing}
\end{table*}

\begin{table*}
\centering\small
\begin{tabular}{p{0.65\linewidth} r}
\toprule
\textbf{Scoring Dimension} & \textbf{Points} \\
\midrule
\textbf{Teaching Objectives (Clarity, Suitability)} & 20 \\
\quad Clear articulation of learning goals & 10 \\
\quad Alignment with standards and student level & 10 \\
\cmidrule{1-2}
\textbf{Content Accuracy and Relevance} & 20 \\
\quad Alignment with subject knowledge & 10 \\
\quad Support for stated objectives & 10 \\
\cmidrule{1-2}
\textbf{Teaching Methods (Variety, Fit)} & 20 \\
\quad Use of diverse, learner-centered strategies & 10 \\
\quad Appropriateness for content and learners & 10 \\
\cmidrule{1-2}
\textbf{Instructional Flow and Coherence} & 20 \\
\quad Logical sequencing of activities & 10 \\
\quad Complete stages (intro, practice, summary) & 10 \\
\cmidrule{1-2}
\textbf{Resource Use} & 10 \\
\quad Richness and suitability of teaching aids & 5 \\
\quad Match with teaching needs & 5 \\
\cmidrule{1-2}
\textbf{Assessment Design} & 10 \\
\quad Diverse and valid evaluation strategies & 5 \\
\quad Process and outcome balance & 5 \\
\cmidrule{1-2}
\textbf{Innovation and Feasibility} & 10 \\
\quad Novelty and classroom applicability & 5 \\
\quad Realistic timing and resource constraints & 5 \\
\midrule
\textbf{Total} & \textbf{100} \\
\bottomrule
\end{tabular}
\caption{GPT-based scoring rubric for Teaching Design tasks.}
\label{tab:gpt-teaching-rubric}
\end{table*}

\end{document}